\documentclass[journal]{IEEEtran}
\usepackage{amsmath,amsfonts}
\usepackage{algorithmic}
\usepackage{algorithm}
\usepackage{array}
\usepackage{textcomp}
\usepackage{stfloats}
\usepackage{url}
\usepackage{verbatim}
\usepackage{graphicx}
\usepackage{cite}
\usepackage{bm}
\usepackage{caption}
\usepackage{subfigure}
\usepackage{booktabs}
\usepackage{threeparttable}
\usepackage{multirow}
\usepackage{enumitem}
\usepackage{xcolor}
\usepackage{mathtools}
\newcommand{\review}[1]{\textcolor{black}{#1}}

\begin{document}

\title{Learning Spatio-Temporal Dynamics for Trajectory Recovery via Time-Aware Transformer}

\author{Tian Sun, Yuqi Chen, Baihua Zheng, Weiwei Sun
\thanks{This research is supported in part by the National Natural Science Foundation of China under grant 62172107 and
the National Key Research and Development Program of China under grant 2018YFB0505000. 

Tian Sun, Yuqi Chen and Weiwei Sun 
are with School of Computer Science, Fudan University, Shanghai, 200433, China,
and also with Shanghai Key Laboratory of Data Science, Fudan university, Shanghai, 200433, China,
and also with Shanghai Institute of Intelligent Electronics \& Systems, Shanghai, 200433, China
(e-mail: tsun23@m.fudan.edu.cn; chenyuqi21@m.fudan.edu.cn;  wwsun@fudan.edu.cn. Corresponding author: Weiwei Sun)

Baihua Zheng is with the School of Computing and Information
Systems, Singapore Management University, Singapore, 188065 (e-mail:
bhzheng@smu.edu.sg).}
}

\markboth{IEEE TRANSACTIONS ON INTELLIGENT TRANSPORTATION SYSTEMS}{}


\maketitle

\begin{abstract}
In real-world applications, GPS trajectories often suffer from low sampling rates, with large and irregular intervals between consecutive GPS points. This sparse characteristic presents challenges for their direct use in GPS-based systems. This paper addresses the task of map-constrained trajectory recovery, aiming to enhance trajectory sampling rates of GPS trajectories. Previous studies commonly adopt a sequence-to-sequence framework, where an encoder captures the trajectory patterns and a decoder reconstructs the target trajectory. Within this framework, effectively representing the road network and extracting relevant trajectory features are crucial for overall performance. Despite advancements in these models, they fail to fully leverage the complex spatio-temporal dynamics present in both the trajectory and the road network. 

\review{To overcome these limitations, we categorize the spatio-temporal dynamics of trajectory data into two distinct aspects: spatial-temporal traffic dynamics and trajectory dynamics. Furthermore, We propose TedTrajRec, a novel method for trajectory recovery. To capture spatio-temporal traffic dynamics, we introduce PD-GNN, which models periodic patterns and learns topologically aware dynamics concurrently for each road segment. For spatio-temporal trajectory dynamics, we present TedFormer, a time-aware Transformer that incorporates temporal dynamics for each GPS location by integrating closed-form neural ordinary differential equations into the attention mechanism. 
This allows TedFormer to effectively handle irregularly sampled data.
Extensive experiments on three real-world datasets demonstrate the superior performance of TedTrajRec. The code is publicly available at \url{https://github.com/ysygMhdxw/TEDTrajRec/}.}
\end{abstract}

\section{Introduction}\label{introduction}

The rapid advancement of Location-Based Services (LBS) has made GPS trajectories a crucial data format with various applications in Intelligent Transportation Systems (ITS)~\cite{ren2021mtrajrec, wang2020combining, wei2020particle, veres2019deep}, including trajectory similarity computation~\cite{yao2022trajgat, zhang2020trajectory, yang2021t3s} and traffic prediction~\cite{zhang2018deeptravel, yi2020automated, zheng2020hybrid, liu2020dynamic}.
These applications require a large volume of high-quality trajectories to achieve optimal performance. 
\review{Despite the growing use of real-time GPS tracking systems, particularly in urban environments and services like online car-hailing, low sampling rates remain a significant challenge in certain situations~\cite{ren2021mtrajrec, yuan2010interactive}. Factors such as low-battery modes or power-saving settings on mobile devices, along with GPS interference in urban canyons or areas with high electromagnetic interference, contribute to these reduced sampling rates, as highlighted in studies on mobile data and GPS-dependent systems~\cite{li2016knowledge, liu2022deepgps, ng2021robust}. Additionally, the sparse temporal and spatial granularity of call detail records (CDR) and mobile phone signaling data (MSD) collected in urban areas further exacerbates this issue~\cite{huang2023accurate, zhao2022identifying}.}

\review{Typically, these trajectories are first map-matched to a road network using methods such as Hidden Markov Models (HMM)~\cite{song2012quick} or other map-matching algorithms~\cite{zhao2019deepmm} before being utilized in downstream applications. Table~\ref{tab:hmm} illustrates the relationship between the sampling rate of GPS trajectories and the accuracy of map-matching using the HMM algorithm~\cite{song2012quick}. As shown in the table, the accuracy of the HMM algorithm drops dramatically as the sampling intervals increase. Consequently, low-sampling-rate GPS trajectories severely impact the effectiveness of downstream ITS applications.}
\review{In response to these challenges, recent research has increasingly focused on low-sample rate GPS trajectories, particularly in the areas of trajectory recovery~\cite{ren2021mtrajrec, chen2023rntrajrec, wei2024micro, zhaograph, wang2024towards, cao2023f} and map matching for such trajectories~\cite{huang2023accurate, jiang2023l2mm}.}

\begin{figure}[t]
    \centering
    \includegraphics[width=\linewidth]{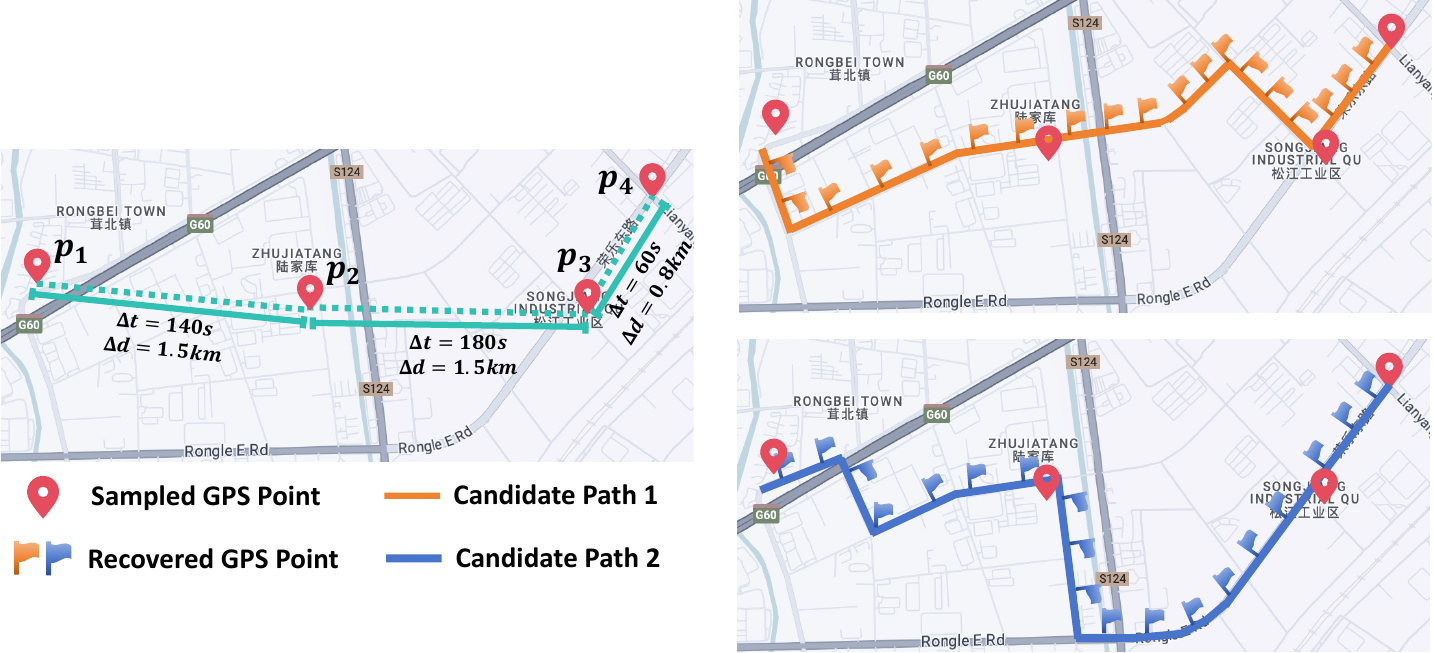}
    \caption{Illustration of Trajectory Recovery Task with Irregularly-Sampled GPS Points.
    }
    \label{fig:trajectory_recovery}
\end{figure}

In this paper, we address the task of trajectory recovery, a foundational challenge in working with low-sample-rate GPS trajectory. As illustrated in Figure~\ref{fig:trajectory_recovery}, such trajectories are characterized by large, irregular time gaps between consecutive GPS points, leading to significant uncertainty and a lack of detailed information on travel behavior and purpose. Specifically, as depicted on the right side of Figure~\ref{fig:trajectory_recovery}, two distinct travel paths can correspond to the same set of sampled input points, highlighting the ambiguity in these datasets.

\begin{table}[t]
\caption{\review{The accuracy of map matching using the Hidden Markov Model (HMM) algorithm w.r.t. the sampling interval of GPS trajectory data on the Shanghai dataset.}}
    \centering
    \resizebox{\columnwidth}{!}{%
    \begin{tabular}{c|ccccc}
    \toprule
         Interval (unit: second) & 10 & 20 & 40 & 80 & 160 \\
    \midrule
         Map Matching Accuracy & 0.9301 &0.8889&0.7791&0.5731&0.3490  \\ 
         
    \bottomrule
    \end{tabular}}
    \label{tab:hmm}
\end{table}

Studies in this area can be broadly categorized into two approaches based on whether the road network structure is incorporated: network-free trajectory recovery~\cite{chen2023teri,elshrif2022network, wang2019deep} and map-constrained trajectory recovery~\cite{ren2021mtrajrec, chen2023rntrajrec, shi2023road}. 
While network-free trajectory recovery offers a more generalized solution~\cite{chen2023teri}, incorporating road network information has shown to significantly improve the performance of trajectory recovery task~\cite{chen2023rntrajrec}. Given the availability of road network data through platforms such as OpenStreetMap, our paper focuses on map-constrained trajectory recovery.

\begin{figure}[t]
  \centering
  \subfigure[\scriptsize{Spatio-temporal Traffic Dynamics}]{\includegraphics[width=0.43\linewidth]{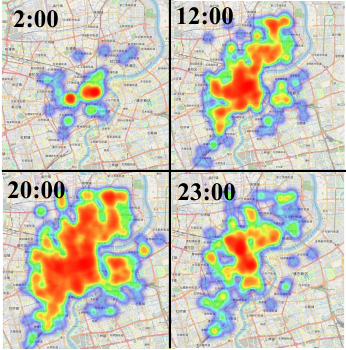}\label{fig:sub1}}
  \hfill 
  \subfigure[\scriptsize{Spatio-temporal Trajectory Dynamics}]{\includegraphics[width=0.54\linewidth]{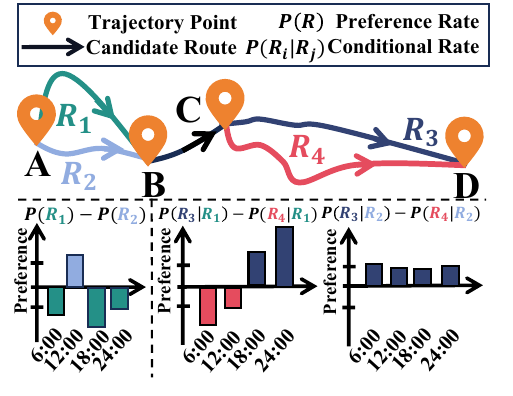}\label{fig:sub2}}
  \caption{Illustration of Spatio-Temporal Dynamics on Road Network and Trajectory.
  }
  \label{fig:dynamic}
\end{figure}


Recovering low-sampling-rate GPS trajectory is challenging. Ren et al.~\cite{ren2021mtrajrec} proposed
the first end-to-end solution for the task, with an encoder to process the input low-sampling-rate trajectories and a decoder to reconstruct missing GPS points. 
Besides, leveraging the powerful ability of the Transformer, Chen et al.~\cite{chen2023rntrajrec} introduced a spatio-temporal Transformer model. 
\review{Wei et al.~\cite{wei2024micro} further proposes a novel graph-based representation model that considers time intervals and distance gaps, achieving state-of-the-art performance.}
\review{However, these approaches often fail to model the critical spatio-temporal dynamic, within the trajectory data and the road network, as illustrated in Figure~\ref{fig:dynamic}.}

\noindent \textbf{(i) Spatio-Temporal Traffic Dynamics:} \review{As illustrated in Figure~\ref{fig:dynamic}(a), which depicts the traffic flow distribution in Shanghai at four different timestamps throughout the day, capturing the time-varying road network structure presents intricate challenges~\cite{DBLP:conf/ijcai/YuYZ18,chen2024maritime, chen2024ship}.} Simply modeling the static topology of the road network~\cite{chen2023rntrajrec} is insufficient for effectively representing these dynamic changes.


\noindent \textbf{(ii) Spatio-Temporal Trajectory Dynamics:} As shown in Figure~\ref{fig:dynamic}(b), due to complex and evolving traffic conditions, decision preferences between routes $R_1$ and $R_2$, as well as $R_3$ and $R_4$, vary both spatially and temporally.
Moreover, because the trajectory recovery is treated as a sequential decision problem~\cite{wu2017modeling}, the decision-making criteria for choosing between $R_3$ and $R_4$ become more intricate and subject to temporal evolution. 
While Jiang et al.~\cite{jiang2023self} and Wei et al.~\cite{wei2024micro} integrate time interval information into trajectory feature extraction, and several other studies~\cite{ren2021mtrajrec, Liang22TrajFormer} include time-related features (e.g., hour embedding), these approaches fail to capture the complex dynamic changes and nuanced patterns across various scenarios. \review{Additionally, the irregularity of GPS trajectory requires careful model design, simply relying on standard Transformers and recurrent neural networks (RNNs) for trajectory encoding~\cite{chen2023rntrajrec, ren2021mtrajrec} may struggle to effectively capture these irregular intervals.}




\review{To comprehensively learn the complex spatio-temporal dynamics within the trajectory and road network, we propose \textbf{TedTrajRec}, a Spatio-\textbf{Te}mporal \textbf{d}ynamics-aware Transformer model for map-constrained \textbf{Traj}ectory \textbf{Rec}overy. }
\review{
To learn spatio-temporal traffic dynamics, we introduce PD-GNN ( \textbf{P}eriodic \textbf{D}ynamic-aware \textbf{G}raph \textbf{N}eural \textbf{N}etwork), a road network representation module that incorporates periodic dynamics. 
PD-GNN captures unique dynamics for each road segment and adopts graph neural networks to enhance spatial patterns.
Additionally, to account for both irregular temporal correlation and spatio-temporal trajectory dynamics in feature extraction, we propose TedFormer (Spatio-\textbf{Te}mporal \textbf{d}ynamics-aware Trans\textbf{former}), a novel time-aware Transformer.}
\review{By integrating closed-form neural ordinary differential equations (Neural ODEs) into the attention mechanism, TedFormer introduces a time-aware attention mechanism that enhances the model's ability to capture the full spatio-temporal dynamics of the data.} 

To summarize, the contributions of this paper are outlined as follows.
\begin{itemize}[leftmargin=15pt]
    \item \review{To the best of our knowledge, this is the first attempt to comprehensively model the spatial-temporal dynamics involved in trajectory recovery. We further categorize these dynamics within trajectory data into two distinct aspects: spatial-temporal traffic dynamics and trajectory dynamics.}
    \item \review{We propose PD-GNN to capture periodic traffic patterns, enhancing road network representation learning for spatio-temporal traffic dynamics.}
    \item \review{We introduce TedFormer to learn spatial-temporal trajectory dynamics, which models GPS trajectories by incorporating closed-form neural ordinary differential equations into the attention mechanism. Notably, Tedformer considers the irregular sampling characteristic of trajectory data.}
    \item We evaluate TedTrajRec on three real-world datasets, showcasing a significant improvement over existing state-of-the-art methods. Additionally, we conduct experiments to illustrate the practical impact of trajectory recovery through two critical upstream tasks.
\end{itemize}

\section{Related Work}

\subsection{Trajectory Recovery.}
Trajectory data faces challenges like low and nonuniform sample rates, along with noisy sample points~\cite{yuan2010interactive}. Trajectory recovery is crucial for reconstructing low-sampling-rate or incomplete trajectory data~\cite{zhang2022enhancing,cao2023f,zhang2023simidtr,deng2023fusing}, ensuring completeness and reliability of trajectory data essential for downstream applications. 
Previous work follows two branches, i.e., network-free trajectory recovery~\cite{elshrif2022network, wang2019deep, hoteit2014estimating} and map-constrained trajectory recovery~\cite{ren2021mtrajrec, chen2023rntrajrec}. In this paper, we focus on the task of map-constrained trajectory recovery. 
\review{
DHTR~\cite{wang2019deep} extends the sequence-to-sequence (seq2seq) generation framework by reconstructing high-density trajectories and using a map-matching algorithm, such as HMM~\cite{newson2009hidden}, for accurate GPS location recovery. Recently, MTrajRec~\cite{ren2021mtrajrec} proposes a seq2seq multi-task learning approach that addresses this task in an end-to-end manner, achieving significant improvement compared to two-stage approaches. Technically, MTrajRec combines constraint mask, attention mechanism, and attribute module to handle coarse grid road network presentations, while overlooking the topology between road segments. Moving forward, RNTrajRec~\cite{chen2023rntrajrec} improves accuracy by incorporating road network topology into a spatio-temporal Transformer architecture. MM-STGED~\cite{wei2024micro} constructs a micro-view of the trajectory based on time differences and distance gaps,  adopting a graph-based approach for trajectory extraction with additional road condition data. 
In this paper, we propose TedTrajRec, which distinguishes itself from existing methods through the proposed road network representation model, PD-GNN, and a novel trajectory recovery model, TedFormer.}

\review{
For the road network representation model, RNTrajRec~\cite{chen2023rntrajrec} adopts a graph-based representation learning approach that focuses solely on the spatial structure of the road network, neglecting critical spatio-temporal traffic dynamics. MM-STGED captures spatio-temporal traffic patterns using only coarse grid-level statistical features, which ignores road network topology 
and fails to learn complex traffic dynamics in large-scale real-life transportation systems. Our proposed PD-GNN learns the temporal dynamics of each road segment inherently, without requiring additional road condition data.}

\review{For the trajectory extraction model, both MTrajRec and DHTR utilize a simple Recurrent Neural Network (RNN) that takes coarse-grained grid information and timestamps as inputs. RNTrajRec employs a spatial-temporal Transformer architecture with topology-aware, fine-grained features. However, their use of RNN and Transformer with only timestamps as input often struggles with irregular input trajectories and is insensitive to variations in time intervals.
MM-STGED considers the time differences and distance gaps within a graph-based approach for trajectory extraction. However, their use of kernel functions, such as the exponential decay function, to model time dynamics remains insufficient in capturing input-dependent dynamic systems~\cite{chen2024contiformer}.
Our method overcomes these limitations by combining dynamic systems modeling with a Transformer architecture. This design allows TedTrajRec to effectively capture input-dependent temporal dynamics and irregularities in GPS input trajectories, providing a significant advantage in feature extraction. 
}


%
\subsection{\review{Modeling Spatio-Temporal Dynamics.}}
\review{Spatio-temporal dynamics, which involve the interplay between spatial and temporal factors, are crucial in various scenarios~\cite{yuan2022learning,yuan2022aligned}, including trajectory prediction~\cite{yu2020spatio, chu2024adaptive} and trajectory generation~\cite{yuan2022activity}.}

\review{For instance, in traffic flow prediction, spatio-temporal dynamics involve both the spatial relationships and temporal sequences of vehicles or individual movements. To model such dynamics, STGODE~\cite{yu2020spatio} and ADM-STNODE~\cite{chu2024adaptive} construct a tensor-based ODE to learn spatial and temporal semantic connections synchronously. However, these approaches overlook the periodicity inherent in traffic flow. Furthermore, their usage of Neural ODEs within structured graph data incurs significant computational costs, limiting their ability to handle the large-scale road network with millions of road segments.}

\review{In the context of activity trajectory generation, spatio-temporal dynamics consist of complex spatio-temporal associations underlying daily activities, often modeled by semantic information (e.g., activity type). ActSTD~\cite{yuan2022activity} integrates an activity decision model with Neural ODEs to capture temporal and spatial dynamics, effectively modeling transition patterns with spatio-temporal dependencies. However, their use of Neural ODEs falls short in capturing long-term sequence dependencies, which are crucial for understanding extended temporal patterns. }


\review{In this paper, we focus on the trajectory recovery for low-sample GPS trajectory. To the best of our knowledge, this marks the first attempt to learn spatio-temporal dynamics for recovering map-constrained GPS trajectory. We formulate two types of dynamics in trajectory recovery, namely spatio-temporal traffic dynamics and spatio-temporal trajectory dynamics. Furthermore, two innovative modules, PD-GNN and TedFormer, are introduced for learning these dynamics respectively, addressing critical issues such as scalability problems, periodic patterns, and long-term sequential dependency issues in existing approaches.}


%

\subsection{Transformer for GPS Trajectory.}
Transformer models have excelled in diverse domains, gaining prominence due to both practical utility and theoretical insights~\cite{DBLP:conf/iclr/ParkK22}.  
In recent years, researchers have been particularly interested in applying the Transformer architecture to GPS trajectory analysis.
For instance, in the field of trajectory similarity computation, Yao et al.~\cite{yao2022trajgat} leverage hierarchical structure modeling techniques and propose a graph-based Transformer model. Chang et al.~\cite{chang2023contrastive} propose a novel dual-feature self-attention-based trajectory backbone encoder. 
Moreover, Liang et al.~\cite{Liang22TrajFormer} propose a continuous point embedding method and a squeezed Transformer model for the trajectory classification task. 
\review{Jiang et al. propose START~\cite{jiang2023self}, a modification to the Transformer architecture by incorporating time interval information as an additional input.}
\review{Ma et al. propose JGRM~\cite{ma2024more}, a joint Transformer model tailored to capture representations of route and GPS trajectories that achieves state-of-the-art performance on various downstream tasks. }
However, these approaches struggle to capture dynamic spatio-temporal contexts, due to the separation of spatial and temporal information.
\review{Thus, we propose a time-aware Transformer, namely TedFormer, that seamlessly incorporates spatio-temporal trajectory dynamics of sequence data by incorporating closed-form Neural ODEs for modeling dynamics of inputs.}

\subsection{Neural Ordinary Differential Equation.}
Neural ODEs\cite{chen2018neural} employ ordinary differential equations (ODEs) to model the continuous dynamics of neural networks, showcasing advancements in irregular time series analysis~\cite{jin2022multivariate, fang2021spatial}, dynamical systems modeling~\cite{bishnoi2022enhancing}, and generative modeling~\cite{yildiz2019ode2vae}. Unlike conventional residual layers, which update the hidden states using transformations at discrete time steps\cite{he2016deep}, Neural ODE captures the continuous dynamics through an ODE specified by a neural network: $\frac{d\textbf{h}(t)}{dt}=f(\textbf{h}(t),t,\theta)$. Here, $t \in \{0,...,T\}$ represents time, $\textbf{h}(t)$ is the hidden state at time step $t$, and $\theta$ denotes neural network parameters.
The model is defined and evaluated using a black-box differential equation solver:
\begin{equation}
  \textbf{h}(0) \cdots \textbf{h}(t_T) =\text{ODESolver}(f, \textbf{h}(0),(t_0,\cdots , t_T)) ~,
\end{equation}
where $\langle t_0, ..., t_T \rangle$ is a sequence of (ir)regularly-sampled timestamps, and $\text{ODESolver}$ is a numerical solver, e.g., Euler method~\cite{euler1794institutiones} or Runge Kuta fourth-order (RK4) method~\cite{runge1895numerische}. 

Applying Neural ODEs to irregular time series modeling is promising for scenarios involving continuous-time dynamics~\cite{kidger2020neural, hasani2021liquid}. In this paper, we propose a novel integration of closed-form ODEs into the Transformer attention mechanism to capture temporal dynamics within trajectory sequence data. This approach leverages Neural ODEs while mitigating efficiency concerns associated with numerical solvers.

\subsection{Road Network Representation Learning.}
Road network representation learning is a powerful tool for spatio-temporal data analysis and management~\cite{kipf2016semi}. Previous studies predominantly treated road networks as directed graphs, with Graph Neural Networks (GNNs) like Graph Isomorphism Networks (GIN)\cite{xu2018powerful}, Graph Attention Networks (GAT)\cite{velivckovic2017graph}, and GATv2\cite{brody2021attentive} gaining prominence in road network representation learning tasks~\cite{chen2023rntrajrec, jiang2023self, li2023trajectory}. Specifically, Li et al.~\cite{li2023trajectory} introduced a partition-based representation learning framework PT2vec, while Zhang et al.~\cite{zhang2023road} proposed a hypergraph to capture high-order relationships. 

Despite these advances, existing approaches often overlook crucial temporal dynamics in road network representation learning, neglecting time-varying information for different road segments. In this work, we surpass current methodologies by incorporating periodic dynamics into road network representation and propose a novel PD-GNN module to comprehensively understand underlying temporal patterns and enhance the temporal dynamics captured in road network representation learning. 

\section{Preliminary}
In this section, we provide an overview of the key definitions and formalize the problem of map-constrained trajectory recovery.

\noindent \textbf{GPS Trajectory.} A GPS trajectory is characterized as a sequence of $m$ GPS points $\rho= \langle p_1,p_2,...,p_m \rangle$, each associated with a timestamp. Here, each GPS point $p_i=(x_i,y_i,t_i)$ is represented by longitude $x_i$ and latitude $y_i$ at time $t_i$. Due to the irregularity of the given trajectory, we define the sampling interval $\epsilon$ as the average time interval between any two consecutive trajectory points. 

\noindent \textbf{Road Network.} A road network is a directed graph $G=(V,E)$, with $V$ representing a set of road segments, and $E$ signifying the set of edges that interconnect these road segments. Each edge $(u,v)$ denotes a direct connection from the road segments $u$ to $v$.

\noindent \textbf{Map Matching.} Due to uncertainties of GPS collection devices, raw trajectory data may not align with the road network. Hence, given road network $G$, the map matching algorithm~\cite{song2012quick} maps each GPS point from the raw trajectory to the corresponding road segment. This process yields a sequence of map-constrained trajectory $\tau=\langle q_1,q_2,...,q_n \rangle$, where each point $q_i=(e_i, r_i, t_i)$ comprises the road segment identifier $e_i$, the position ratio $r_i$ along that segment (e.g., $r_i=0.5$ indicates the midpoint), and the timestamp $t_i$.

Throughout the paper, we denote $\rho$ as the raw GPS trajectory, which is the direct output from GPS devices, whereas the map-constrained GPS trajectory is denoted as the sequence $\tau$.

\noindent \textbf{Trajectory Recovery Task.} As shown in Figure~\ref{fig:trajectory_recovery},  a low-sampling-rate raw GPS trajectory $\rho=\langle p_1, p_2, ..., p_m \rangle$ is characterized by an average sample interval $\epsilon_\rho$, as shown on the left side of the figure, where $m$ represents the input length of the trajectory. The task of map-constrained trajectory recovery is formulated to reconstruct a 
map-constrained GPS trajectory $\tau=\langle q_1, q_2, ..., q_n \rangle$, with a sample interval $\epsilon_\tau$, where $n$ represents the target length of the trajectory. 

\noindent \textit{Remark.} To ensure a lower sampling interval of the recovered trajectory, it is crucial that $n$ is significantly greater than $m$.

\section{Proposed Method}

\begin{figure*}[htbp]
    \centering
    \includegraphics[width=1.0\textwidth]{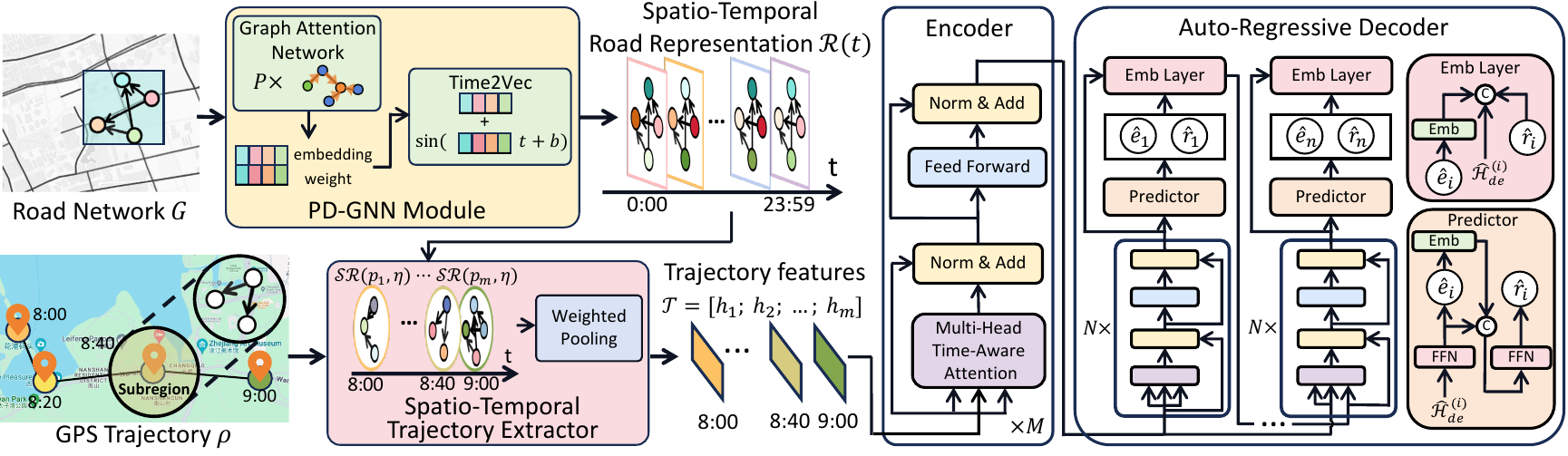}
    \caption{The Architecture of TedTrajRec. TedTrajRec contains \review{i) Feature Extraction, which involves the proposed PD-GNN for learning spatio-temporal traffic dynamics and a trajectory feature extraction module.} \review{ii) Encoder, a time-aware Transformer, namely TedFormer that captures spatio-temporal trajectory dynamics of GPS locations within the attention mechanism.} iii) Auto-Regressive Decoder, a decoder model that employs TedFormer and produces the target trajectory.}    \label{fig:model}
    \vspace{-0.1in}
\end{figure*}

In this section, we introduce TedTrajRec, our innovative approach to trajectory recovery, as illustrated in Figure~\ref{fig:model}.
The key motivation behind TedTrajRec is to leverage the rich spatial and temporal dynamics for accurate trajectory recovery. 
To achieve this, we first present a novel temporal-aware road network representation module, i.e., PD-GNN. This module first models the topological structure of road segments using graph neural networks. It subsequently integrates temporal information into the road network utilizing a time encoding technique~\cite{kazemi2019time2vec}. \review{The output of this module is a road network imbued with spatio-temporal traffic dynamics, with time-varying vector representations for each road segment.} 

To further enhance the learning of dynamic trajectory patterns from the irregularly sampled GPS trajectories, we propose TedFormer, a novel time-aware Transformer architecture. This innovative approach replaces the conventional attention module with a time-aware attention mechanism, designed to jointly model continuous temporal dynamics and spatial information. By integrating a closed-form neural ODE function, TedFormer enables the precise extraction of intricate temporal relationships within the data.

\review{The following sub-sections are organized as follows:}

\begin{itemize}
    \item \review{Sub-section A: Feature Extraction -- In this section, we introduce the novel PD-GNN module, which enhances road network representation by capturing spatio-temporal traffic dynamics. This approach addresses the limitation of existing approaches that focus on either spatial modeling~\cite{chen2023rntrajrec} or statistical temporal patterns~\cite{wei2024micro}. STTExtractor derives trajectory features by leveraging both spatio-temporal road network representations and the surrounding spatial structures of GPS points.}
    \item \review{Sub-section B: Time-Aware Transformer -- Here, we present TedFormer, an innovative module that integrates closed-form Neural ODEs with the attention mechanism. This integration allows TedFormer to effectively model irregularly sampled trajectory data by embedding spatio-temporal trajectory dynamics directly into the attention process.}
    \item \review{Sub-sections C and D: Auto-Regressive Prediction and Loss Function -- In these sections, we follow established methodologies~\cite{ren2021mtrajrec} and adopt an auto-regressive decoder architecture alongside a multi-task learning algorithm to recover missing points in the input trajectory. Notably, TedFormer is also applied within the decoder model, further enhancing its performance.}
\end{itemize}


\subsection{Feature Extraction}

Spatio-temporal information inherent in road network topology plays a crucial role in trajectory feature extraction. To integrate periodic dynamics into road networks and improve trajectory feature extraction, we introduce PD-GNN, a novel periodic dynamic-aware graph neural network. Additionally, leveraging the road network representation, we employ a spatio-temporal trajectory extractor called STTExtractor to capture the interplay between spatial and temporal patterns within trajectory sequences. By synergizing the capabilities of both modules, our proposed method effectively discerns underlying patterns and trends in trajectory data, thereby facilitating enhanced precision in trajectory recovery predictions.

\subsubsection{Spatio-Temporal Road Network Representation Learning}\label{PD-GNN}
To capture the topological structure of the road network and 
model the dependencies between road segments,  
we first adopt the Graph Attention Network (GAT).
Given a road network $G=(V,E)$ and an initial embedding of road segments $S \in \mathbb{R}^{|V| \times d}$, we obtain a high-dimensional road representation vector $\bar{S} \in \mathbb{R}^{|V| \times d}$ by stacking $P$ layers of GATv2~\cite{brody2021attentive} module. Within the module, the attention weights $a_{ij, k}^{(p)}$, representing the importance between road segment $e_i$ and road segment $e_j$ for the $k^{\text{th}}$ attention head at the $p^{\text{th}}$ layer, are calculated as follows:
\begin{eqnarray}
a_{ij, k}^{(p)}=&\frac{\exp \Bigl(\mathrm{a}_k^{(p)^\top} \text{LeakyReLU}\left(\mathbf{\widehat{W}}_k^{(p)}\left[ u_{i}^{(p-1)} \|  u_j^{(p-1)} \right]\right)\Bigr)}{\sum_{o \in \mathcal{N}_i} \exp \Bigl(\mathrm{a}_k^{(p)^\top}\text{LeakyReLU}\left(\mathbf{\widehat{W}}_k^{(p)}\left[ u_{i}^{(p-1)} \|  u_o^{(p-1)}\right]\right)\Bigr)} ~
\label{eq:gat-attn}
\end{eqnarray}
Here, $p \in [1, .., P]$, $u_{i}^{(p)} \in \mathbb{R}^d$ is the embedding vector of road segment $e_i$ at the $p^{\text{th}}$ layer with $u_i^{(0)}=S_i$, $\mathrm{a}_k^{(p)}$ and $\mathbf{\widehat{W}}_k^{(p)}$ are the learnable parameter vectors, \review{$\mathcal{N}_i$ represents the set of neighboring road segments that are at most one-hop away from road segment $e_i$ (by this definition, $e_i \in \mathcal{N}_i$)}, and
$[ \cdot \| \cdot ]$ denotes the concatenation operation.
Subsequently, the multi-head attention mechanism is employed to complete the graph attention network:
\begin{eqnarray}
u_{i}^{(p)}=&\Vert_{k=1}^{K} \text{LeakyReLU}\left(\sum\nolimits_{j \in \mathcal{N}_{i}} a_{ij,k}^{(p)} \mathbf{\widehat{W}}_k^{(p)} u_{j}^{(p-1)}\right)~
\label{eq:gat-h}
\end{eqnarray}
\review{Here, $a_{ij,k}^{(p)}$ represents the attention weight calculated and $K$ is the number of heads. $u_i^{(p)}$ represents the embedding vector of road segment $e_i$ at the $p^{th}$ layer of GATv2 and notation $||_{k=1}^K$ denotes the concatenation operation across the outputs of $K$ attention heads.} The resulting output is obtained by concatenating the elements of the set $\bar{S} = \{u^{(P)}_i \}$. 
\begin{figure}[t]
    \centering
    \includegraphics[width=1.0\linewidth]{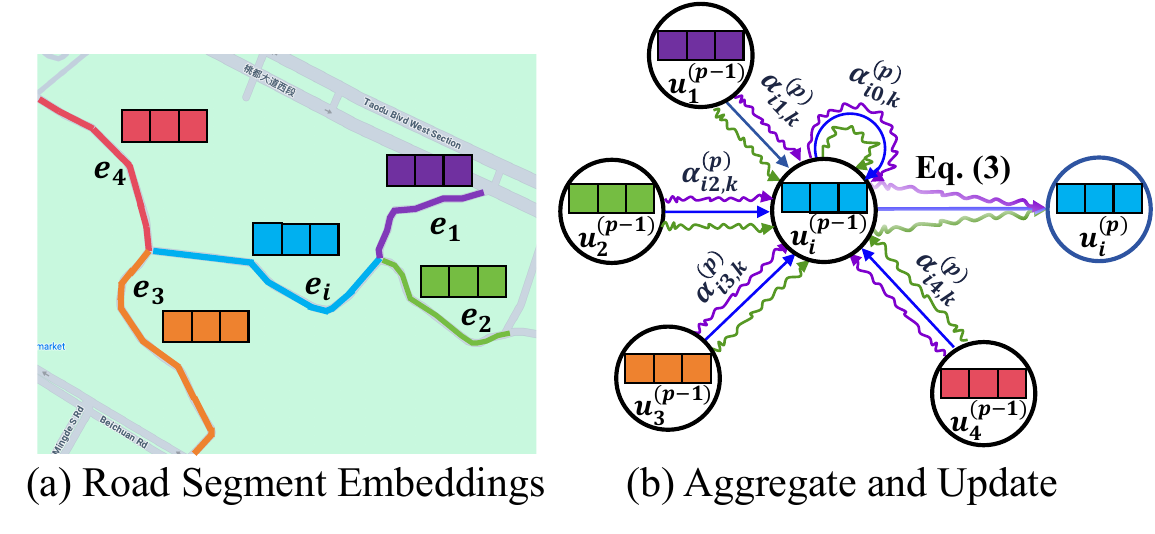}
    \caption{\review{The detailed illustration of the graph attention network on the road network. In sub-figure (a), the central road segment $e_i$, colored in blue, has four neighborhood road segments $e_1, e_2, e_3,$ and $e_4$. Each road segment contains its road segment embeddings. In sub-figure (b), the road segment embedding of $e_i$ is updated by aggregating the embeddings from neighborhood road segments and the embeddings from itself.}}
    \label{fig:gat-v2}
    \vspace{-0.1in}
\end{figure}

\review{The detailed illustration of the process is provided in Figure~\ref{fig:gat-v2}. The central road segment $e_i$ in Figure~\ref{fig:gat-v2}(a), colored in blue, is surrounded by four neighboring road segments, i.e., $\mathcal{N}_i=\{e_1, e_2, e_3, e_4, e_i\}$. In each layer of the graph attention network, the model first aggregates embeddings from the neighboring nodes (including itself) and calculates attention scores, as specified in Eq.~\eqref{eq:gat-attn}. Then, as per Eq.~\eqref{eq:gat-h}, the embedding of the central segment $e_i$ is partitioned into $K$ heads, which are subsequently updated through an aggregation process. The final results are  obtained by concatenating the outputs from each head. This procedure is consistently applied across all road segments in the network, ensuring that each segment effectively leverages information from its neighbors.}

After obtaining the road segment embeddings of $\bar{S}$, the hidden vector is further transformed by two linear transformations to obtain two distinct hidden vectors: $\hat{S} \in \mathbb{R}^{|V| \times d}$ and $\Omega \in \mathbb{R}^{|V| \times d}$, representing the initial embedding and the temporal-related weighted matrix, respectively.
While the road network representation incorporates the topological structure, it fails to capture the nuanced temporal dynamics specific to individual road segments. To remedy this limitation, we attempt to encode temporal information into the road network representation. Specifically, for a timestamp $t \in [0:00, 23:59]$ in the unit of minute, we calculate a temporal embedding \textbf{t2v} as follows:
\begin{eqnarray}
\label{eq:time2vec}
\textbf{t2v}(t; \omega)[i]=
\left\{
\begin{aligned}
           & \omega_i t+b, & \text{if } i=0;\\
          &  \sin (\omega_i t+b), & \text{if } 1\leq i \leq d.\\
\end{aligned}
\right.
\end{eqnarray}
Here, $\omega \in \mathbb{R}^d$ represents the learnable weight parameter, and the bias term $b$ introduces additional flexibility to the transformation. The sine function is employed as a periodic activation function, which adeptly captures the periodical temporal patterns inherent in the time-aware embeddings of road segments.

By combining the extracted road network information with an embedding enriched by temporal data, PD-GNN achieves a continuous-time road network representation for each road segment. \review{Specifically, the fused embedding $X^{road}_{e_i}(t)$ is obtained by summing the temporal-related encoding $\textbf{t2v}(t;\Omega_i)$ and the spatial feature $\hat{S}_i$, which is the temporal-related hidden vectors of road segment $e_i$, i.e.,}
\begin{equation}
    \review{X^{road}_{e_i}(t) := \mathcal{R}([\Omega_i,\hat{S}_i], t) = \textbf{t2v}(t; \Omega_i)+\hat{S}_i ~.}
    \label{Eq:road embedding}
\end{equation}

\review{Since each road segment $e_i$ learns a separate weight $\Omega_i$, PD-GNN integrates both spatial features and temporal embeddings, thus capturing the variation in traffic flow over time for each individual road segment.  This approach does not require additional information beyond GPS point data, as the temporal dynamics are embedded directly within the GPS data and the road network's topological structure.}

\review{In summary, PD-GNN presents a robust approach for analyzing the spatio-temporal dynamics of road networks and generates the final continuous-time road network representation $X^{road}(t)$.}

\vspace{3pt}
\noindent \textbf{Discussion: How PD-GNN Model handles Spatio-Temporal Traffic Dynamics.} \review{In PD-GNN, we leverage a dynamical system approach where the state of the system (in this case, the road segments on the road network) evolves over time. Mathematically, a dynamical system is one in which a function describes the time dependence of a point, i.e., $y=f(x,t)$. As defined in Eq. \eqref{eq: subgraph}, $\mathcal{R}([\Omega_i,\hat{S}_i], t)$ models the time evolving function of the joint vector of $[\Omega_i,\hat{S}_i]$, i.e., the combined embedding for segment $e_i$. Since $[\Omega_i,\hat{S}_i]$ is ubiquitous for each road segment, PD-GNN successfully captures the spatio-temporal traffic dynamics.} 

\subsubsection{Spatio-Temporal Trajectory Extractor - STTExtractor} \label{st traj ex}

Given the final continuous-time road network representation $\mathcal{R}(t)$ from PD-GNN, we subsequently propose STTExtractor, which further utilizes the surroundings of trajectory points on the road network to transform the trajectory sequences to representation vectors.
Specifically, for a sampled GPS point $p_i \in \rho$, we delineate a subregion $\mathcal{SR}(p_i, \eta)$ with a radius $\eta$ centered at  $p_i$. 
Additionally, since the closer road segments should exert a more significant contribution to the trajectory point's representation, we introduce an exponential function to model the influence of road segments from the subregion on the given GPS point $p_i$. Thus, the trajectory point's representation  
 $h_i \in \mathbb{R}^{d}$ is defined as:
\begin{eqnarray}
h_i= \frac{ \sum_{e_j \in \mathcal{SR}(p_i, \eta) } \exp(-\text{dist}^2(e_j,p_i)/\gamma^2) \times X^{road}_{e_j}(t_i)}{\sum_{e_j \in \mathcal{SR}(p_i, \eta) } \exp(-\text{dist}^2(e_j,p_i)/\gamma^2)} ~
\label{eq: subgraph}
\end{eqnarray}
Here, $\text{dist}(e_j,p_i)$ denotes the spherical distance between the GPS point $p_i$ and its projection onto the road segment $e_j$, $X^{road}_{e_j}(t_i)$ is the embedding of the road segment $e_j$ at timestamp $t_i$ obtained from PD-GNN, and $\gamma$ represents a hyper-parameter. 

Finally, we obtain the final trajectory sequence representation $\mathcal{T}=[h_1; h_2; ...; h_m]$ for each $p_i \in \rho$, which combines the road network representation of the graph structure with the spatial and temporal information derived from the trajectory sequence.

\subsection{Time-Aware Transformer}

Efficiently capturing temporal dynamics is pivotal for an in-depth understanding and effective modeling of time-dependent data.
However, when applying Transformer~\cite{vaswani2017attention} to model irregularly sampled GPS trajectories, Transformers often overlook crucial relative time information~\cite{ma2022non}. 
Existing approaches either focus solely on temporal information~\cite{shukla2021multi} or treat spatial and temporal correlations separately~\cite{jiang2023self}, failing to fully capture their interdependencies.
To overcome this limitation, we propose a novel time-aware Transformer architecture that integrates a closed-form dynamic system within the attention mechanism. In our proposed approach, the traditional self-attention mechanism is replaced with the proposed time-aware attention mechanism, which effectively captures both spatial and temporal dependencies in GPS trajectory data.


\subsubsection{Multi-head Time-Aware Attention Mechanism}\label{multi-head tw attn}

\begin{figure}[t]
    \centering
    \includegraphics[width=0.9\linewidth]{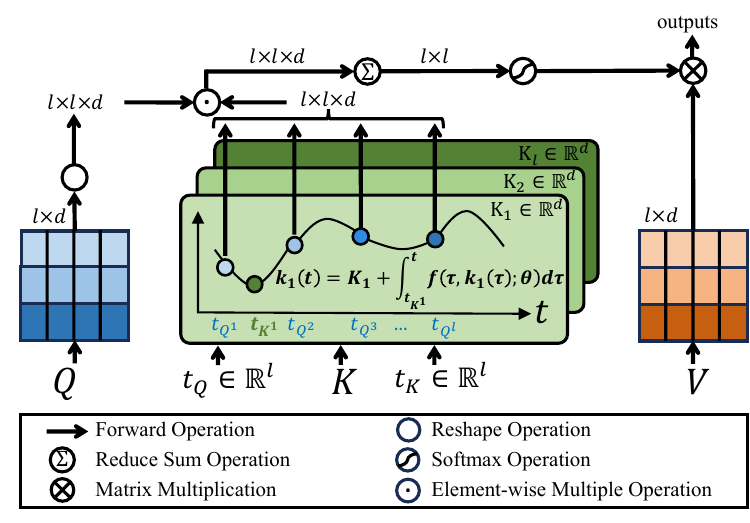}
    \caption{The Overall Architecture of the Multi-head Time-Aware Attention Mechanism in TedFormer.}
    \label{fig:tw-attn}
    \vspace{-0.1in}
\end{figure}

The procedure of integrating neural ODEs within the multi-head time-aware attention mechanism is depicted in Figure \ref{fig:tw-attn}. For simplification, we elaborate on the process under the self-attention scenario.
Given queries $Q \in \mathbb{R}^{l \times d}$, keys $K \in \mathbb{R}^{l \times d}$, and values $V \in \mathbb{R}^{l \times d}$, where $l$ denotes the length of the sequence (or trajectory),
the mechanism first formulates neural ODEs for each sample point in $K$:
\begin{equation}
    \textbf{k}_i(t)=K_i + \int_{t_{K^i}}^{t} f(\tau, \textbf{k}_i(\tau); \theta) d \tau ~, 
\label{eq:ki}
\end{equation}
where $t_{K^i}$ indicates the timestamp for the $i^{\text{th}}$ sample point in $K$, and the initial state of the ODE is given by $\textbf{k}_i(t_{K^i}) = K_i$. Here, $f(\cdot): \mathbb{R}^{d+1} \rightarrow  \mathbb{R}^{d}$ is a neural network governing the change in the ordinary differential function over time. The process is:
\begin{equation}
    \frac{d \mathbf{k}_i(t)}{d t} = f(t, \textbf{k}_i(t); \theta) ~.
\end{equation}

However, solving ODEs with an ODE solver is computationally intensive~\cite{chen2018neural} and may lead to unbounded behavior~\cite{hasani2021liquid}. To address this, Hasani et al.~\cite{hasani2022closed} propose a closed-form continuous-depth network (CfC), which serves as an approximation of liquid time-constant dynamics (LTCs)~\cite{hasani2021liquid}. The closed-form formulation is:
\begin{equation}
\begin{aligned}
\mathbf{x}(t)& =\sigma\left(-\xi_1(x_0; \theta_1) t\right) \odot \xi_2(x_0; \theta_2) \\
& +\left[1-\sigma\left(-\xi_1(x_0; \theta_1) t\right) \right] \odot \xi_3(x_0; \theta_3) ~. 
\end{aligned}
\label{eq:cfc}
\end{equation}
Here, $\xi_1(\cdot)$, $\xi_2(\cdot)$ and $\xi_3(\cdot)$ are neural networks parametrized by $\theta_1$, $\theta_2$ and $\theta_3$, respectively and $\sigma$ denotes the \textit{sigmoid} operation. $x_0$ is the initial state, e.g., $\mathbf{x}(0)$.


Accordingly, the closed-form formulation of $\textbf{k}_i(t)$, initially defined in Eq.~\eqref{eq:ki}, can be defined as:
\begin{equation}
\begin{aligned}
\mathbf{k}_i(t)& =\sigma\left(-\xi_1(K_i; \theta_1) (t-t_{K^i})\right) \odot \xi_2(K_i; \theta_2) \\
& + \left[1-\sigma\left(-\xi_1(K_i; \theta_1) (t-t_{K^i}) \right)\right] \odot \xi_3(K_i; \theta_3) ~. 
\end{aligned}
\label{eq:cfc2}
\end{equation}

Throughout the experiments, $\xi_1(\cdot)$ is defined as:
\begin{eqnarray}
\xi_1(x) = \text{Lecun} (W_1 x^\top + b_1),
\label{eq:cfccell}
\end{eqnarray}
where $\theta_1=\{ W_1, b_1\}$, and \textit{Lecun}~\cite{lecun1989generalization} is a modified \textit{tanh} activation function following  \cite{hasani2022closed}. A similar approach is adopted to define $\xi_2(\cdot)$ and $\xi_3(\cdot)$.

Subsequently, a spatio-temporal key matrix, $K^{ST} \in \mathbb{R}^{l \times l \times d}$, is obtained by considering $t_Q$:
\begin{equation}
\small
\operatorname{K^{ST}}(K, t_Q, t_K)=\left[\begin{array}{cccc}
\textbf{k}_1(t_{Q^1}) & \textbf{k}_2(t_{Q^1}) & \cdots & \textbf{k}_l(t_{Q^1}) \\
\textbf{k}_1(t_{Q^2}) & \textbf{k}_2(t_{Q^2}) & \cdots & \textbf{k}_l(t_{Q^2}) \\
\vdots & \vdots & \ddots & \vdots \\
\textbf{k}_1(t_{Q^l}) & \textbf{k}_2(t_{Q^l}) & \cdots & \textbf{k}_l(t_{Q^l})
\end{array}\right] ~. \label{eq:kk}
\end{equation}

Next, $\text{TedAttn}(\cdot) \in \mathbb{R}^{l \times l}$ is calculated as:
\begin{equation}
\small
\text{TedAttn} (Q, K, t_Q, t_K)[i, j]=  Q_i \cdot \operatorname{K^{ST}}(K, t_Q, t_K)[i, j]^\top ~.
\label{eq: tedattn}
\end{equation}

Intuitively, TedAttn accounts for the temporal dynamics of each sampled point within the trajectory and captures temporal correlations. Furthermore, by introducing $t_Q$ and $t_K$ in Eq.~\eqref{eq:kk}, TedAttn effectively models temporal information.

To enhance the attention mechanism~\cite{vaswani2017attention}, we employ multi-head attention, which facilitates concurrent focus on various aspects of the input. Finally, the multi-head time-aware attention is calculated as: 
\begin{equation}
\small
\begin{aligned}
\text{TA-MHA} (Q, K, V, t_Q, t_K)  
 =\text{Concat}\left(\operatorname{head}_1, \ldots, \operatorname{head}_H\right) W^O ~, 
\end{aligned}
\end{equation}
where $\operatorname{head}_{h}$ is defined as:
\begin{equation}
\small
\operatorname{head}_{h} = \text{SM} \left( \frac{\text{TedAttn}\left(Q W_{Q, h}^\top, K W_{K, h}^\top, t_Q, t_K \right)}{\sqrt{d}}  \right) V W_{V, h}^\top ,
\end{equation}
Here, $W^O$, $W_{Q, h}$, $W_{K, h}$ and $W_{V, h}$ are parameter matrices, $h \in [1,H]$, $H$ is the head number, and SM stands for the softmax operation. 

Additionally, we define: 
\begin{equation}
\small
\begin{aligned}
    \text{TA-MHA-S}(\mathcal{X};t_\mathcal{X}) & = \text{TA-MHA}(\mathcal{X}, \mathcal{X}, \mathcal{X}, t_\mathcal{X}, t_\mathcal{X}) ~, \\
    \text{TA-MHA-C}(\mathcal{X}_{d}, \mathcal{X}_{e};t_{d},t_{e}) & = \text{TA-MHA}(\mathcal{X}_{d}, \mathcal{X}_{e}, \mathcal{X}_{e}, t_{d}, t_{e}) ~.
\end{aligned}
\end{equation}
for self-attention and cross-attention, respectively. Overall, by incorporating a closed-form dynamic system, TedAttn adeptly captures the intrinsic spatio-temporal interdependencies.

\subsubsection{TedFormer Structure}

Despite the widespread adoption of Transformers~\cite{vaswani2017attention} in various research fields, their extension for modeling temporal information is underexplored. We propose TedFormer that directly builds upon the original implementation of vanilla Transformer while replacing the multi-head attention with the proposed time-aware attention mechanism. Assuming we adopt $M$ layers of TedFormer, we formally characterize the TedFormer for the $l^{\text{th}}$ encoder layer as $\mathcal{H}_{\text{en}}^l=\text{EncoderLayer}(\mathcal{H}_{\text{en}}^{l-1};t_{\text{en}})$: 
\begin{eqnarray}
\begin{aligned}
\widetilde{\mathcal{H}}_{\text{en}}^{l}& = \text{LayerNorm}(\text{TA-MHA-S}(\mathcal{H}_{\text{en}}^{l-1};t_{\text{en}})+\mathcal{H}_{\text{en}}^{l-1}) ~, \\
\mathcal{H}_{\text{en}}^{l} &= \text{LayerNorm}(\text{FFN}(\widetilde{\mathcal{H}}_{\text{en}}^{l})+\widetilde{\mathcal{H}}_{\text{en}}^{l}) ~.
\end{aligned}
\end{eqnarray}
Here, $t_{\text{en}}$ represents the time information and $\text{FFN}$ is a fully connected network with \textit{ReLU} activation function.

\vspace{10pt}
\noindent \textbf{Discussion: TedFormer Models Irregular Temporal Correlation.}
\review{Transformers are well-suited for capturing temporal correlations, but they typically assume uniform spacing between tokens. In contrast, TedFormer models irregular temporal correlations by calculating the relationship between tokens sampled at arbitrary time points $t_i$ and $t_j$~\cite{ma2022non}.}
\review{Specifically, the correlation function takes four variables: $Q_i$, $K_j$ (the embeddings of the $i$-th and $j$-th tokens w.r.t. queries and keys), and $t_{Q^i}$, $t_{K^i}$ (the irregularly-sampled timestamps). In a standard Transformer architecture, the correlation function is defined as $\text{corr}(Q_i, K_j;t_{Q^i},t_{K^j})=Q_i \cdot K_j^\top$.}
\review{In TedFormer, however, the correlation is adjusted to account for the time difference, as shown in Eq.~\eqref{eq: tedattn}.}
\review{This formulation allows TedFormer to handle irregular temporal correlation effectively.}

\vspace{10pt}
\noindent \textbf{Discussion: TedFormer Learns Spatial-Temporal Trajectory Dynamics.}
\review{TedFormer also accounts for the temporal dynamics of each sampled point within the trajectory. As shown in Eq.~\eqref{eq:ki}, each sampled point in the trajectory exhibits temporal dynamics, which are modeled using neural ordinary differential equations. Mathematically, this is expressed as $y=f(x,t)$, where the function $f$ describes the evolution of a spatial point $x$ along time. Additionally, TedFormer incorporates spatio-temporal patterns extracted by the STTExtractor, which models the trajectory's spatial and temporal dependencies. The dynamic systems $[\textbf{k}_1(\cdot), ..., \textbf{k}_N(\cdot)]$ are thus capable of handling spatio-temporal dynamics of the trajectory.}

\subsection{Auto-Regressive Prediction} 

Consider a trajectory $\rho=\langle p_1, ..., p_m \rangle$, where $p_i=(x_i, y_i, t_i)$. First, we extract spatio-temporal features, i.e., $\mathcal{T}$. Next, we define $\mathcal{H}_{\text{en}}^{0}=\mathcal{T} + PE$ and the corresponding timestamps for the encoder, i.e., $t_{\text{en}}=\{ t_1, ..., t_m\}$. PE indicates the positional embedding~\cite{vaswani2017attention}. Subsequently, we forward them into $M$ encoder layers to obtain the output, i.e. $\mathcal{H}_{\text{en}}^{M}=\text{Encoder}(\mathcal{H}_{\text{en}}^{0}; t_{\text{en}})$.

We adopt an auto-regressive decoder diagram. 
Given the timestamp sequence to predict, i.e., $t_{\text{de}}=\{\hat{t}_1, ..., \hat{t}_n\}$, the initial hidden-state vectors at timestamp $0$ are calculated as $\mathcal{H}_{\text{de}}^{(0)}=\frac{1}{m} \sum_{i=1}^{m} \mathcal{H}_{\text{en}, i}^{M}$. 

Next, the hidden-state vectors at timestamp $\hat{t}_i$ are given by:
\begin{equation}
\begin{aligned}
\widehat{\mathcal{H}}_{\text{de}}^{(i)}&=\text{Decoder}(\mathcal{H}_{\text{de}}^{(i-1)}, \mathcal{H}_{\text{en}}^{M}; \hat{t}_i, t_{\text{en}}) ~, \\
\hat{e}_i, \hat{r}_i &= \text{Predictor}(\widehat{\mathcal{H}}_{\text{de}}^{(i)}) ~, \\
\mathcal{H}_{\text{de}}^{(i)} &= \text{Emb}(\hat{e}_i, \hat{r}_i, \widehat{\mathcal{H}}_{\text{de}}^{(i)}) ~.
\end{aligned}
\end{equation}
Here, $\mathcal{H}_{\text{de}}^{(i)}$ represents the decoder output at timestamp $\hat{t}_i$. The decoder module takes in the inputs and forwards them through $N$ layers of decoder layers. $\text{Predictor}$ takes in the hidden-state vectors and outputs the predicted road segment id $\hat{e}_i$ and position ratio $\hat{r}_i$, as detailed in Section~\ref{predictor}. Subsequently, we define $\text{Emb}$ as an embedding layer to obtain the input for the next timestamp, which is elaborated in Section \ref{emb}.

\subsubsection{Multi-Task Trajectory Recovery Predictor}\label{predictor}

Aimed at recovering the low-sampling-rate trajectory to a map-constrained trajectory with a target sample interval $\epsilon_\tau$, the multi-task trajectory recovery predictor takes the $\widehat{\mathcal{H}}_{\text{de}}^{(i)}$ as input and obtains the prediction at timestamp $\hat{t}_i$, i.e., $\hat{e}_i$ and $\hat{r}_i$. 
To account for the typical maximum error in GPS data collection, previous work~\cite{ren2021mtrajrec} defines a constraint mask $\mathcal{C} \in \mathbb{R}^{n \times |V|}$ which serves as a prior probability that associates $p_i$ with road segments. 
%
%
Thus, we predict the $j^{\text{\text{th}}}$ road segment as follow:
\begin{equation}
P_\Theta(\hat{e}_j | \widehat{\mathcal{H}}_{\text{de}}^{(i)})=\frac{\exp(\widehat{\mathcal{H}}_{\text{de}}^{(i)} \cdot \textbf{w}_{\hat{e}_j}^\top ) \odot \mathcal{C}_{i,\hat{e}_j}}{\sum_{e' \in V} \exp(\widehat{\mathcal{H}}_{\text{de}}^{(i)} \cdot \textbf{w}_{e'}^\top ) \odot \mathcal{C}_{i, e'}} ~.
  \label{constraint probability function}
\end{equation}
Here, $\textbf{w} \in \mathbb{R}^{|V| \times d}$ represents learnable matrix weight, and $\widehat{\mathcal{H}}_{\text{de}}^{(i)}$ represents the input hidden layer vector obtained from multi-head time-aware attention mechanism presented in Section \ref{multi-head tw attn}. $\odot$ represents the element-wise multiplication. $P_\Theta(\hat{e}_j | \widehat{\mathcal{H}}_{\text{de}}^{(i)})$ represents the predicted output result of the model for road segment $\hat{e}_j$ at the $i^{\text{th}}$ decoding timestamp. The predicted road segment $\hat{e}_i$ is determined by selecting the one with the maximum posterior probability.

Given the road network representation $X^{road}(t) $ generated from Section~\ref{PD-GNN} and the road segment $\hat{e}_i$, we calculate the predicted position ratio of point $\hat{r}_i$ by:
\begin{eqnarray}
   \hat{r}_i=R\left( \left[\widehat{\mathcal{H}}_{\text{de}}^{(i)} \|  X_{\hat{e}_i}^{road}(\hat{t}_i) \right]\right) ~.
\end{eqnarray}
Here, $R(x): \mathbb{R}^{2d} \rightarrow \mathbb{R}$ refers to a neural network used for predicting the position ratio $\hat{r}_i$ of the vehicle on road segment $\hat{e}_i$.

\subsubsection{Embedding Layer} \label{emb}
To obtain the input for the next timestamp, the embedding layer takes into account both the previous prediction and the previous hidden-state vector:
\begin{equation}
\text{Emb}(\hat{e}_i, \hat{r}_i, \widehat{\mathcal{H}}_{\text{de}}^{(i)})=  [X_{\hat{e}_i}^{road}(\hat{t}_i)\| \hat{r}_i \| \widehat{\mathcal{H}}_{\text{de}}^{(i)}] \cdot W^{Emb} + b^{Emb} ~.
\end{equation}
Here, $W^{Emb} \in \mathbb{R}^{(2d+1) \times d}$ and $b^{Emb} \in \mathbb{R}^{d}$ transform the features to a $d$-dimensional space, and $X^{road}_{\hat{e}_i}(\hat{t}_i)$ represents the road segment state given the prediction $\hat{e}_i$ at timestamp $\hat{t}_i$.

\subsection{Loss Function}
We employ the cross entropy loss function for road segment ID prediction~\cite{ren2021mtrajrec, chen2023rntrajrec}, as follows:
\begin{eqnarray}
    \mathcal{L}_{id}=- \sum\nolimits_{\rho, \tau \in \mathcal{D} }  \sum\nolimits_{j=1}^{n} \log \left(
     P_\Theta(e_i|\widehat{\mathcal{H}}_{\text{de}}^{(j)})\right)~.\label{eq:loss}
\end{eqnarray}
Here, $\rho$ and $\tau$ present the input low-sampling-rate trajectory and the map-constrained trajectory with the target sample interval $\epsilon_\tau$, respectively, and $\mathcal{D}$ indicates the training dataset.

Furthermore, the loss function for the recovered position ratio of the trajectory point is defined as:
\begin{eqnarray}
      \mathcal{L}_{ratio}=\sum\nolimits_{\rho,\tau  \in \mathcal{D} }\sum\nolimits_{j=1}^{n}\left(r_j-\hat{r}_j\right)^2~.
\end{eqnarray}

In summary, the overall loss function of the final model is a combination of two individual loss functions, i.e.,
\begin{eqnarray}
\mathcal{L}_{total}=\mathcal{L}_{id}+\lambda\mathcal{L}_{ratio} ~. \label{eq:overall_loss}
\end{eqnarray}
Here, the parameter $\lambda$ is a tunable parameter. The entire algorithm is trained using the backpropagation through time algorithm (BPTT). 

\section{Experiment}

In this section, we validate the efficacy of TedTrajRec in trajectory recovery through comprehensive experiments on three real-world datasets. 
\review{These experiments include an overall comparison study between TedTrajRec and 13 baselines, a robustness test of TedTrajRec and existing map-constrained trajectory recovery approaches for input trajectories with different sampling intervals, and an efficiency study to rigorously examine the computational efficiency of TedTrajRec and existing map-constrained trajectory recovery methods.
Additionally, we perform ablation studies to evaluate the contributions of TedTrajRec's key modules and a parameter sensitivity analysis. Furthermore, we present a case study to demonstrate TedTrajRec's effectiveness and adaptability in one specific scenario. 
Lastly, an upstream analysis highlights the practical impact of trajectory recovery on two pivotal upstream tasks.}

\subsection{Experiment Setting}

\subsubsection{Datasets}
Our experiments are based on three substantial trajectory datasets derived from real-life scenarios in diverse urban environments: Shanghai, Chengdu, and Porto~\cite{kaggle-dataset}. 
We selected the trajectory in urban areas from Chengdu and Porto as training data. To evaluate the scalability of our model, we curated the Shanghai-L dataset, which encompasses both suburban and central urban regions in Shanghai. It is important to highlight that the Shanghai-L dataset covers a training area exceeding 700 km$^2$, equivalent to the land areas of many metropolitan cities such as Singapore (734.3 km$^2$), New York City (778.18 km$^2$), and Mumbai (603.4 km$^2$).
The road networks utilized in our study were extracted from OpenStreetMap (\url{www.openstreetmap.org}).
Detailed information about these datasets is presented in Table~\ref{tab:datasets}.

To ensure a standardized approach, we followed the dataset pre-processing pipeline in~\cite{chen2023rntrajrec}, selecting roughly $150,000$ random trajectories from each dataset for this experimental study. These datasets were then split into training, validation, and testing sets with a ratio of 7:2:1. 

Trajectory recovery studied in this paper aims to reconstruct high-quality map-constrained trajectories from low-sampling-rate trajectories.
To create ground truth data, we utilized the HMM algorithm~\cite{song2012quick} and linear interpolation techniques to generate map-constrained trajectories with the target sample interval $\epsilon_\tau$. 
Furthermore, to generate low-sampling-rate trajectories with irregular time intervals, we adopt a random downsampling method for each trajectory within the dataset, i.e., a given probability drops each sample point in the trajectory.
%
For each dataset, the average time interval of the low-sampling-rate trajectories is set to be 8 or 16 times (i.e., $\epsilon_\rho=\epsilon_\tau \times 8$ (or $16$)) greater than the sampling interval of the original trajectories slated for reconstruction. 
For instance, for the setting of $\epsilon_\rho=\epsilon_\tau \times 8$, each sample point within the high-quality map-constrained trajectories is dropped with a probability of $87.5\%$.

\begin{table}[t]
\centering
\caption{Statistics of Datasets}
\label{tab:datasets}
 \resizebox{\columnwidth}{!}{%
\begin{tabular}{@{}cccc@{}}
\toprule
Dataset                                   & Shanghai-L & Chengdu   & Porto   \\ \midrule
Trajectory collected duration & Apr 2015 & Nov 2016 & Jul 2013-Mar 2014 \\
\# Trajectories                            & 2,694,958       & 8,302,421 & 999,082 \\
\# Road Segment                            & 34,986          & 8,781     & 12,613  \\
Size of training area ($\operatorname{km}^2$) & $23.0 \times 30.8$ & $8.3 \times 8.3$ & $6.8 \times 7.2$ \\
Average time span per trajectory ($s$) & 699.57          & 868.86    & 783.14  \\
Sample interval $\epsilon_\tau$ ($s$)       & 10            & 12      & 15    \\ 
\midrule
\review{\# Total trajectories for model} &\review{149,835}&\review{151,256}&\review{157,516}\\
\review{\# Training set trajectories} &\review{104,962}& \review{105,879}&\review{110,261}\\
\review{\# Validation set trajectories} &\review{14,994}& \review{15,125}&\review{15,751}\\
\review{\# Testing set trajectories} &\review{29,989}&\review{30,252}&\review{31,504}\\
\bottomrule
\end{tabular}
}
\end{table}

\begin{table*}[htbp]
\renewcommand\tabcolsep{3.3pt} 
\caption{Performance evaluation on trajectory recovery task. A higher value for the Recall, Precision, F1 Score, and Accuracy metrics indicates a better performance, while a lower value for the MAE and RMSE metrics indicates a better performance. Notably, MAE and RMSE are in meters.}
\vspace{-0.15in}
\begin{center}
\resizebox{\textwidth}{!}{%
\begin{tabular}{c|c|cccccc|cccccc}
\toprule
\multicolumn{2}{c|}{\multirow{2}*{Method}}&\multicolumn{6}{c|}{Chengdu ($\epsilon_\rho=\epsilon_\tau*8$)}&\multicolumn{6}{c}{Chengdu ($\epsilon_\rho=\epsilon_\tau*16$)} \\
\multicolumn{2}{c|}{~} & Recall & Precision & F1 Score & Accuracy & MAE & RMSE & Recall & Precision & F1 Score & Accuracy & MAE & RMSE  \\
\midrule
\multirow{4}{*}{\hspace{0.05cm}\rotatebox{90}{Two-Stage}\hspace{0.05cm}} & Linear + HMM & 0.6597 & 0.6166 & 0.6351 & 0.4916 & 358.24 & 594.32 & 0.4821 & 0.4379 & 0.4564 & 0.2858 & 525.96 & 760.47 \\
& TrImpute + HMM &0.6644 &0.6306 & 0.6450 &0.5083&381.72& 620.64 & 0.4981 & 0.4671 & 0.4797& 0.2933&535.95 &777.61 \\
& DHTR + HMM & 0.6385 & 0.7149 & 0.6714 & 0.5501 & 252.31 & 435.17 & 0.5080 & 0.6930 & 0.5821 & 0.4130 & 325.14 & 511.62 \\
& TERI + HMM & 0.7430 & 0.6803& 0.7076& 0.5468& 252.16&448.33 & 0.7066&0.6593&0.6792&0.4294&311.45&501.59 \\
\midrule
\multirow{10}{*}{\hspace{0.05cm}\rotatebox{90}{End-to-End}\hspace{0.05cm}} 
& NeuTraj + Decoder & 0.7795 & 0.8731 & 0.8214 & 0.6407 & 145.48 & 238.28 & 0.6771 & 0.8466 & 0.7492 & 0.5182 & 213.43 & 324.37 \\ 
& GTS + Decoder & 0.7746 & 0.8788 & 0.8212 & 0.6399 & 145.87 & 241.44 & 0.6739 & 0.8615 & 0.7530 & 0.5188 & 212.10 & 321.76 \\
& TrajGAT + Decoder & 0.7778 & 0.8733 & 0.8207 & 0.6414 & 146.70 &242.59 &
  0.6793 & 0.8511 & 0.7523 & 0.5283 &206.92 & 317.52\\
& TrajFormer+ Decoder &0.7726 &0.8790&0.8203&0.6506&139.80&230.94     &  0.6816 & 0.8500&0.7534&0.5166&211.94&327.09    \\
& \review{START + Decoder} &\review{0.7827}&\review{0.8824}&\review{0.8275}&\review{\underline{0.6680}}&\review{\underline{132.29}}&\review{223.09}&\review{ {0.6926}}&\review{0.8544}&\review{0.7620}&\review{0.5442}&\review{193.66}&\review{ {298.37}}\\
& \review{JGRM + Decoder} &\review{0.7810}&\review{\textbf{0.8959}}&\review{\underline{0.8325}}&\review{0.6594}&\review{134.40}&\review{222.92} & 
\review{\underline{0.6936}}&\review{\underline{0.8753}}&\review{\underline{0.7709}}&\review{\underline{0.5510}}&\review{\underline{186.90}}&\review{\underline{292.26}}\\
& MTrajRec & 0.7701 & 0.8694 & 0.8144 & 0.6360 & 150.60 & 247.21 & 0.6714 & 0.8473 & 0.7460 & 0.5152 & 216.37 & 327.37 \\
& RNTrajRec & 0.7813 & {0.8851} & 0.8278 & 0.6628 & 133.08 & \underline{221.74} & {0.6889} & {0.8712} & {0.7663} & 0.5378 & 198.75 &306.19 \\ 
& \review{MM-STGED} & \review{\underline{0.7849}} & \review{0.8822} & \review{{0.8287}} & \review{{0.6655}} & \review{134.12} & \review{228.75}  & \review{0.6794} & \review{0.8479} & \review{0.7511} & \review{{0.5464}} & \review{{192.68}} & \review{{300.60}}\\
& \textbf{TedTrajRec} & \textbf{0.8247} & \underline{0.8936} & \textbf{0.8561} & \textbf{0.7301} & \textbf{108.33} & \textbf{192.22} & \textbf{0.7191} & \textbf{0.8789} & \textbf{0.7882} & \textbf{0.5943} & \textbf{171.19} & \textbf{274.68} \\ 
\midrule
\multicolumn{2}{c|}{\textbf{Improvements}} & \review{\textbf{+5.07\%}} & \review{-0.26\%} & \review{\textbf{+2.83\%}} & \textbf{\review{+9.30\%}} & \textbf{\review{+18.11\%}} & \review{\textbf{+13.31\%} } & \textbf{\review{+3.67\%} }&\textbf{\review{+0.41\%} }& \textbf{\review{+2.24\%}} & \textbf{\review{+7.86\%}}& \textbf{\review{+9.18\%} }& \textbf{\review{+6.40\%}}
\\
\bottomrule
\end{tabular}}

\vspace{5pt}
\hspace{0em}
\resizebox{\textwidth}{!}{%
\begin{tabular}{c|c|cccccc|cccccc}
\toprule
\multicolumn{2}{c|}{\multirow{2}*{Method}}&\multicolumn{6}{c|}{Porto ($\epsilon_\rho=\epsilon_\tau*8$)}&\multicolumn{6}{c}{Shanghai-L ($\epsilon_\rho=\epsilon_\tau*8$)} \\
\multicolumn{2}{c|}{~} & Recall & Precision & F1 Score & Accuracy & MAE & RMSE & Recall & Precision & F1 Score & Accuracy & MAE & RMSE  \\
\midrule
\multirow{4}{*}{\hspace{0.05cm}\rotatebox{90}{Two-Stage}\hspace{0.05cm}} & Linear + HMM & 0.5837 & 0.5473 & 0.5629 & 0.3624 & 175.00 & 284.16 & 0.6055 & 0.5633 & 0.5801 & 0.3825 & 383.25 & 555.68  \\
& TrImpute + HMM & 0.5832 & 0.5520 & 0.5650 & 0.3639& 176.51& 289.07 & \underline{0.7649} &0.7304 & 0.7443& 0.5992& 243.59& 403.69 \\
& DHTR + HMM & 0.5578 & 0.6837 & 0.6118 & 0.4250 & {104.41} & {168.83} & 0.5144 & 0.6533 & 0.5696 & 0.3974 & 308.72 & 454.87  \\
& TERI + HMM & 0.6255& 0.5592& 0.5877&0.3370&133.11&212.47  &0.7239 &0.5885&0.6438&0.4995&249.88&464.37 \\
\midrule
\multirow{10}{*}{\hspace{0.05cm}\rotatebox{90}{End-to-End}\hspace{0.05cm}} 
& NeuTraj + Decoder & 0.6709 & 0.7878 & 0.7219 & 0.5075 & 107.12 & 159.03 & 0.7435 & 0.7907 & 0.7606 & 0.5878 & 201.59 & 311.78 \\ 
& GTS + Decoder & 0.6706 & 0.8006 & 0.7272 & 0.5139 & 102.87 & 152.05 & 0.7527 & 0.8101 & 0.7749 & 0.6041 & 183.11 & 282.96 \\ 
& TrajGAT + Decoder &   0.6741 & 0.7823 & 0.7216 & 0.5026 &109.99 & 162.89 &  0.7388 & 0.8022 & 0.7636 & 0.5884 & 200.27 & 310.89 
\\
& TrajFormer+ Decoder & 0.6663
&\textbf{0.8145}&0.7305&0.5313&96.86&143.93 & 0.7383&0.8402&0.7811&0.6240&172.46&269.57 \\
& \review{START + Decoder} & \review{0.6789}&\review{0.8107}&\review{0.7365}&\review{0.5332}&\review{95.52}&\review{141.92} & \review{0.7408}&\review{0.8726}&\review{0.7969}&\review{0.6370}&\review{161.57}&\review{254.62}\\
& \review{JGRM + Decoder} &\review{0.6717}&\review{\underline{0.8142}}&\review{0.7338}&\review{0.5315}&\review{94.80}&\review{140.32}
& \review{{0.7633}}&\review{\underline{0.8795}}&\review{\underline{0.8135}}&\review{\underline{0.6570}}&\review{\underline{153.35}}&\review{\underline{239.30}}\\
&  MTrajRec & 0.6742 & 0.7833 & 0.7219 & 0.4989 & 110.72 & 163.45 & 0.7140 & 0.7677 & 0.7334 & 0.5561 & 219.29 & 332.19 \\
& RNTrajRec & 0.6833 & 0.8106 & 0.7392 & 0.5365 & 94.15 & 139.94 & 0.7475 & {0.8562} & 0.7938 & {0.6392} & 162.67 & 257.06 \\
& \review{MM-STGED} &  \review{\underline{0.6917}} &  \review{0.8116} &  \review{\underline{0.7402}} &  \review{\underline{0.5376}} &  \review{\underline{93.02}} &  \review{\underline{136.95}} & \review{{0.7512}} & \review{0.8524} & \review{{0.7940}} & \review{0.6388} & \review{{157.93}} & \review{{252.29}} \\
& \textbf{TedTrajRec} & \textbf{0.7097} & 0.8120 & \textbf{0.7554} & \textbf{0.5852} & \textbf{79.49} & \textbf{123.86} & \textbf{0.8057} & \textbf{0.8944} & \textbf{0.8447} & \textbf{0.7321} & \textbf{129.06} & \textbf{225.08} \\
\midrule
\multicolumn{2}{c|}{\textbf{Improvements}} & \review{\textbf{+2.60\%}} & \review{-0.31\%} &\review{\textbf{+2.05\%}} & \review{\textbf{+8.85\%}} & \review{\textbf{+14.55\%}} & \review{\textbf{+10.57\%}} & 
\textbf{\review{+5.33\%}} & \textbf{\review{+1.69\%}} & \textbf{\review{+3.84\%}} & \textbf{\review{+11.43\%}} & \textbf{\review{+18.82\%}} & \textbf{\review{+5.43\%}} \\
\bottomrule
\end{tabular}}
\label{tab:performace}
\end{center}
\end{table*}

\subsubsection{Evaluation Metrics}

Given a predicted trajectory $\hat{\tau}= \langle \hat{q}_{1}, \hat{q}_{2},$ $..., \hat{q}_n \rangle$ and its ground truth trajectory $\tau= \langle q_1, q_2,..., q_n \rangle$, where $\hat{q}_i=(\hat{e}_i, \hat{r}_i, \hat{t}_i)$ and $q_i=(e_i, r_i, \hat{t}_i)$, we evaluate the performance of map-constrained trajectory recovery.
Our evaluation focuses on both the accuracy of recovered road segments and the distance error of location inferences.

\noindent
\textbf{Accuracy.}
The accuracy metric evaluates the model's ability to predict the correct road segments aligned with the target road segments, i.e., $\text{Accuracy}(\tau, \hat{\tau})=\frac{1}{n}\sum^{n}_{i=1}1\{e_i=\hat{e_i}\}$.

\noindent
\textbf{MAE \& RMSE.} 
Mean Absolute Error (MAE) and Root Mean Square Error (RMSE) assess the model's performance by measuring the shortest path distance between the predicted map-constrained GPS points generated by the model and the ground truth~\cite{ren2021mtrajrec}. Specifically, $\text{MAE}(\tau, \hat{\tau})=\frac{1}{n}\sum^{n}_{i=1}\text{dist}(q_i,\hat{q}_i)$ and $\text{RMSE}(\tau, \hat{\tau})=\sqrt{\frac{1}{n}\sum^{n}_{i=1}\text{dist}(q_i,\hat{q}_i)^2}$.  

\noindent
\textbf{Recall \& Precision \& F1 Score.}
Given the predicted road segment sequence $E_{\hat{\tau}}=\langle \hat{e}_1, \hat{e}_2, ..., \hat{e}_n\rangle$, and the target road segment sequence $E_{\tau}=\langle e_1, e_2, ..., e_n\rangle$, we calculate recall, prediction, and F1 scores to evaluate the match between them.  
Specifically, $\text{Recall}\left(\tau, \hat{\tau}\right)=\frac{|E_{\tau} \cap E_{\hat{\tau}}|}{|E_{\tau}|}$, $\text{Precision}\left(\tau, \hat{\tau}\right)=\frac{|E_{\tau} \cap E_{\hat{\tau}}|}{|E_{\hat{\tau}}|}$, and  $\text{F1}\left(\tau, \hat{\tau}\right)=\frac{2 \times \text{Recall}\left(\tau, \hat{\tau}\right) \times \text{Precision}\left(\tau, \hat{\tau}\right)}{\text{Recall}\left(\tau, \hat{\tau}\right) + \text{Precision}\left(\tau, \hat{\tau}\right)}$. 

\subsubsection{Parameter Settings}
All experiments were conducted on a machine with an AMD Ryzen 9 5950X 16-core CPU and a 24GB NVIDIA GeForce RTX 3090 GPU. The implementation of TedTrajRec and all baselines utilized the PyTorch framework~\cite {paszke2019pytorch}. Training across all models spanned 30 epochs, employing the Adam optimizer with a batch size of 64 and a learning rate of $10^{-3}$. To accommodate memory limitations, the dimension of hidden-state vectors was set to 256, maintaining consistency across all models on the three datasets. 
Following ~\cite{chen2023rntrajrec}, we set $\eta$ (used to define subregions around GPS points) and $\gamma$ (used in Eq.~\eqref{eq: subgraph}) to $400$ meters and $30$ meters, respectively. 
Both the number $P$ of GATv2 layers and the number $M$ of encoder layers are fixed at $2$, and the number $N$ of decoder layers is fixed at $1$. Furthermore, we set $\lambda$ used in the overall loss function defined in Eq.~\eqref{eq:overall_loss} to $1$.

\subsubsection{Compared Models}
To assess the effectiveness of TedTrajRec, we selected 13 models as baselines. These baselines can be categorized into two-stage approaches and end-to-end approaches.

\noindent \textbf{Two-Stage Approach.} These methods employ either spatio-temporal imputation techniques or deep learning methods to reconstruct GPS trajectories in the absence of a road network. Subsequently, they utilize the HMM algorithm~\cite{song2012quick} to generate map-constrained trajectories. Specifically, these methods are: 

\begin{itemize}[leftmargin=15pt]
    \item \textbf{Linear}~\cite{hoteit2014estimating} that
employs a linear interpolation to recover the missing GPS points.
    \item \textbf{TrImpute}~\cite{elshrif2022network} that leverages index construction to generate candidate points. 
    \item \textbf{DHTR}~\cite{wang2019deep} that is a seq2seq model with Kalman filter~\cite{kalman1960new} as a feature extractor for linear dynamic systems. 
    \item \review{\textbf{TERI}~\cite{chen2023teri}, the state-of-the-art approach for network-free trajectory recovery, that combines a Transformer encoder and transition pattern learning.}
\end{itemize}

\noindent \textbf{End-to-End Approach.} 
These methods employ deep learning methods that directly predict map-constrained trajectories. 
For a thorough assessment of TedTrajRec, we additionally compare it with recent trajectory similarity and trajectory classification tasks for feature extraction, as well as, recent approaches for trajectory representation. For these models, we adopt the encoder structures from these models, while utilizing the decoder from Ren et al., ~\cite{ren2021mtrajrec} to recover missing trajectory points. Specifically, we compare TedTrajRec with the following trajectory encoders:

\begin{itemize}
    \item \textbf{NeuTraj}~\cite{yao2019computing} that leverages a spatial attention memory augmented LSTM~\cite{hochreiter1997long} module for trajectory encoding.
    \item \textbf{GTS}~\cite{han2021graph} that combines graph neural network and LSTM models and captures spatial network and sequence information jointly. 
    \item \textbf{TrajGAT}~\cite{yao2022trajgat} that employs a PR quadtree~\cite{samet1988overview} to build the hierarchical structure of the spatial area and proposes a graph Transformer to encode the trajectory.
    \item \textbf{TrajFormer}~\cite{Liang22TrajFormer} that adopts a Transformer-based structure by employing a continuous point embedding and a squeezed Transformer block.
    \review{\item \textbf{START}~\cite{jiang2023self} that captures temporal regularities and travel semantics for trajectory representation, achieving superior performance on a variety of downstream trajectory-based tasks.}
    \review{\item \textbf{JGRM}~\cite{ma2024more}, the state-of-the-art approach for trajectory representation, that jointly models GPS and route representation with Transformer architecture.}
\end{itemize}

Finally, we compare TedTrajRec with existing approaches tailored for map-constrained trajectory recovery:

\begin{itemize}[leftmargin=15pt]
    \item \textbf{MTrajRec}~\cite{ren2021mtrajrec} that introduces a multi-task seq2seq learning architecture. 
    \item \textbf{RNTrajRec}~\cite{chen2023rntrajrec} that proposes a novel spatio-temporal Transformer encoder for processing GPS trajectories.
    \review{
    \item \textbf{MM-STGED}~\cite{wei2024micro}, the state-of-the-art trajectory recovery approach, that leverages a graph-based method for trajectory feature extraction and constructs a micro-view of the trajectory that takes into account both time differences and distances.}
\end{itemize}

\subsection{Performance Evaluation}

We detail the experimental outcomes in Table~\ref{tab:performace}, revealing TedTrajRec's remarkable superiority over baseline models across three real-world datasets. 
Numbers in bold highlight the best performance, while numbers that are underlined indicate the second-best performance. The reported improvements refer to the ratio of improvements achieved by TedTrajRec compared to the best performance achieved by other baseline methods.

In particular, two-stage models exhibit inferior performance, highlighting the importance of learning how to represent the road network. 
Among these four comparative two-stage models, 
TrImpute outperforms the others on the Shanghai-L dataset because it effectively utilizes neighborhood information to identify candidate points in a large dataset.
Additionally, TERI shows a 13.03\% increase in F1 Score on the Shanghai-L dataset compared to the DHTR model, due to its ability to capture long-term dependencies in trajectory sequences.  

\review{Moreover, our comparison also includes models specifically designed for trajectory similarity and classification tasks. By leveraging the powerful representation capabilities of the Transformer architecture, TrajFormer outperforms other approaches such as NeuTraj, GTS, and TrajGAT, which use RNNs or graph-based Transformers, on both the Porto and Shanghai-L datasets. }

\review{Finally, we conduct a comprehensive comparison of TedTrajRec against strong baselines in trajectory representation (START~\cite{jiang2023self} and JGRM~\cite{ma2024more}) and map-constrained trajectory recovery (RNTrajRec~\cite{chen2023rntrajrec} and MM-STGED~\cite{wei2024micro}). As shown in the table, the MAE and RMSE metrics of the four models mentioned above significantly outperform those of previous baselines. Additionally, START+Decoder and JGRM+Decoder outperform RNTrajRec. In Chengdu ($\epsilon_\rho = \epsilon_\tau * 8)$ dataset, START+Decoder exceeds RNTrajRec in accuracy and MAE metrics by 0.78\% and 0.59\% respectively. This improvement can be attributed to its integration of temporal information within the Transformer, leading to enhanced performance. On the other hand, JGRM+Decoder also outperforms RNTrajRec in terms of precision and F1 score, demonstrating that combining route and GPS information within a Transformer model yields superior results. Meanwhile, MM-STGED slightly outperforms both START+Decoder and JGRM+Decoder, and consistently outperforms RNTrajRec on the MSE and RMSE metrics. 
Last but not the least, our newly proposed TedTrajRec achieves the best performance in almost all metrics across different datasets. Even in the only two cases where TedTrajRec does not perform the best, the difference between TedTrajRec and the best performer is nearly negligible. Furthermore, TedTrajRec achieves an overall improvement of 9.37\% in terms of the accuracy metric and 13.70\% in terms of the MAE metric compared to the best baseline model. These findings highlight the significance of learning spatio-temporal dynamics for trajectory recovery.}



\subsection{Robustness of Varied Input Sample Intervals}

\review{To demonstrate the robustness of TedTrajRec across various scenarios, we conducted extensive experiments using low-sampling-rate trajectories with varying sampling intervals. For the Chengdu dataset, the target trajectory sampling interval was fixed at 12 seconds, while for the Porto dataset, it was set to 15 seconds. The average sampling intervals of the input trajectories ranged from 24 to 192 seconds for Chengdu, and from 30 to 240 seconds for Porto. In other words, the ratio of sampling rate of target trajectory to that of input trajectory $\epsilon_\rho/{\epsilon_\tau}$ varies in the range of $[2, 4, 8, 16]$.
}

\begin{figure*}[htbp]
    \centering
    \includegraphics[width=1.0\textwidth]{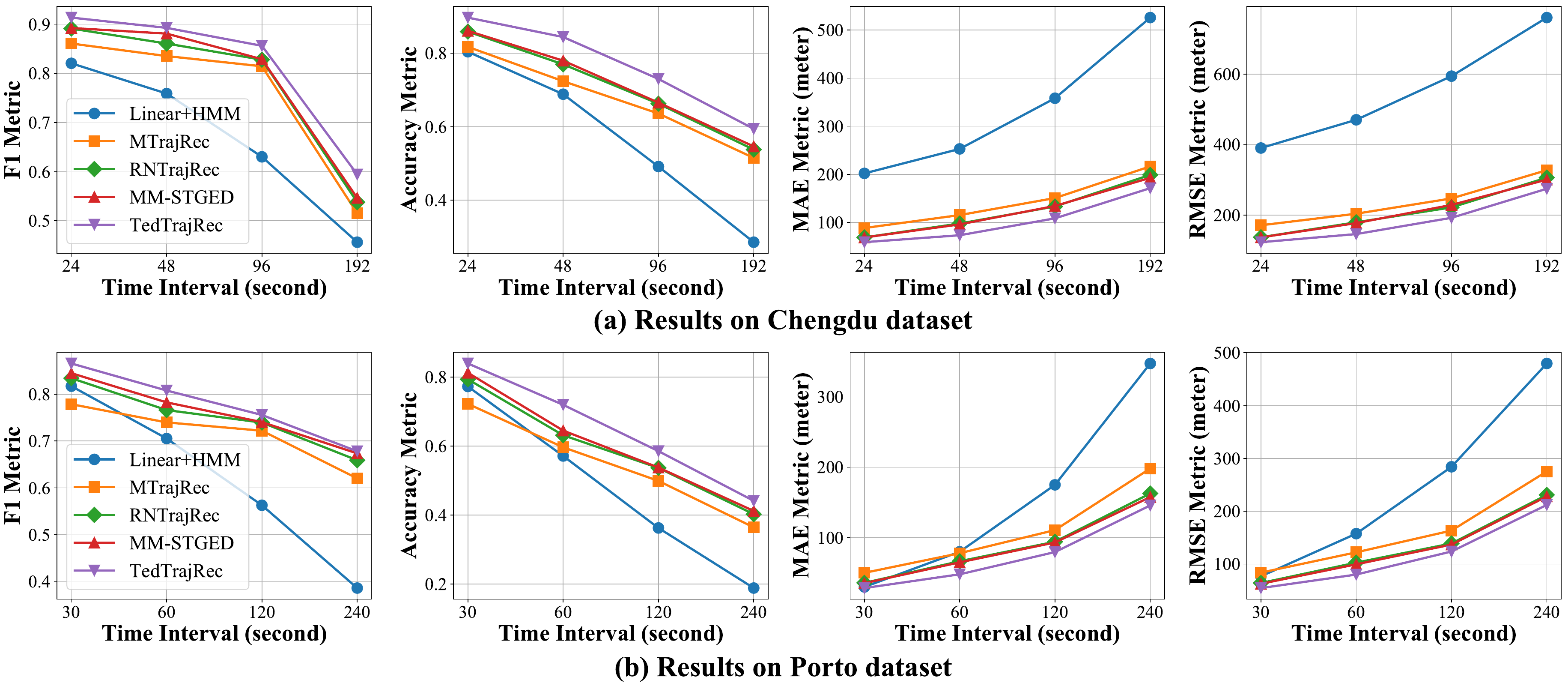}
    \caption{\review{Performance evaluation of trajectory recovery task on input trajectories with varying sampling intervals, using the Chengdu and Porto Datasets. The sampling interval of the target trajectories is fixed at 12 seconds for the Chengdu dataset and 15 seconds for the Porto dataset.
    }
    }    \label{fig:multiratio}
\end{figure*}

\review{
We compare TedTrajRec with several strong baselines designed for map-constrained trajectory recovery, including MTrajRec, RNTrajRec, and MM-STGED. Additionally, we incorporate Linear+HMM into the study, as HMM is particularly effective when the sampling interval is small. As shown in Figure~\ref{fig:multiratio}, TedTrajRec consistently outperforms other methods across all sampling intervals. Notably, the performance gap between TedTrajRec and the Linear+HMM approach widens as the sampling interval increases, suggesting that deep learning-based trajectory recovery methods are particularly effective for larger intervals. Specifically, on the Chengdu dataset, the accuracy gap between Linear+HMM and TedTrajRec is 11.6\% when the sampling interval is 24 seconds, but it grows to 108\% when the interval increases to 192 seconds. Additionally, the gap between TedTrajRec and MM-STGED, the state-of-the-art approach for map-constrained trajectory recovery, is 4.19\% at a sample interval of 24 seconds, increasing to 8.76\% at a sample interval of 192 seconds.}

\subsection{Efficiency Study}

\begin{table*}[t]
\renewcommand\tabcolsep{3.3pt} 
\caption{\review{Efficiency study of TedFormer on Chengdu dataset ($\epsilon_\rho=\epsilon_\tau * 8$).}}
\vspace{-0.1in}
\begin{center}
\resizebox{0.9\linewidth}{!}{%
\begin{tabular}{c|c|c|c|c|ccc}
\toprule
Model & Params (MB) & MACs ($\times 10^{10}$) & Max Memory (GB) & Epoch Time (minute) & F1  & Accuracy & RMSE \\
\midrule
MTrajRec&8.56&4.03&1.26&17&0.8144&0.6360&247.21\\
RNTrajRec&14.57&98.47&11.44&27&0.8278&0.6628&\underline{221.74}\\
MM-STGED& 16.98 & {95.24}&{8.69} & {21} & \underline{0.8287} & \underline{0.6655} &{228.75}\\
\midrule
TedTrajRec& 5.92&37.61&8.77&23&\textbf{0.8561}&\textbf{0.7301}&\textbf{192.22}\\
\bottomrule
\end{tabular}}
\label{tab: effi}
\end{center}
\vspace{-0.15in}
\end{table*}

In this paper, we propose TedFormer for the trajectory recovery task. TedFormer seamlessly integrates closed-form neural ordinary differential equation functions with the attention mechanism, making it well-suited for irregularly sampled GPS trajectories that exhibit long-term dependencies. \review{To examine the effectiveness of TedTrajRec, we first consider the time complexity of the encoder and decoder models against various approaches proposed for map-constrained trajectory recovery, with results shown in Table~\ref{tab:timecost}. }Specifically, the time complexity for the encoder refers to the complexity for $\mathcal{H}_{\text{en}}^{M}=\text{Encoder}(\mathcal{H}_{\text{en}}^{0}; t_{\text{en}})$, while decoder time complexity refers to the complexity for recovering \textit{one} point, i.e., $\widehat{\mathcal{H}}_{\text{de}}^{(i)}=\text{Decoder}(\mathcal{H}_{\text{de}}^{(i-1)}, \mathcal{H}_{\text{en}}^{M}; \hat{t}_i, t_{\text{en}})$. 
\review{Typically, when differential equations are used in the context of neural networks, they are solved using ODE solvers~\cite{chen2018neural}, with computational complexity generally dependent on the accuracy of the solver. However, our approach uses a closed-form differential equation, as defined in Eq.~\eqref{eq:cfc} and Eq.~\eqref{eq:cfc2}, eliminating the need for ODE solvers in the implementation. This reduces the time complexity of calculating $\mathbf{k}_i(t)$ in Eq.~\eqref{eq:cfc2} to $O(d^2)$, and $\operatorname{K^{ST}}(K, t_Q, t_K)$ in Eq.~\eqref{eq:kk} to $O(m^2 \times d^2)$.} 

For the decoder model, given the auto-regressive decoder paradigm, the total complexity for recovering one trajectory is $n \times$ the decoder complexity for one point. \review{As shown in the table, TedTrajRec demonstrates the same complexity for both the decoder and the decoder model compared to the state-of-the-art method, i.e., MM-STGED. Additionally, it is important to note that $\|G\|$ is larger than $m$, making TedTrajRec's encoder comparable to RNTrajRec.}

\begin{table}[t]
    \centering
    \caption{\review{Time complexity analysis of various models for map-constrained trajectory recovery in end-to-end approaches. $m$ stands for the length of the input trajectories and $d$ stands for the dimension of the hidden-state vectors. $\|G\|$ stands for the size of the sub-regions, i.e., $\|G\|= \sum_{i=1}^{m} \| \mathcal{SR}(p_i, \eta)\|$.
    }}
\resizebox{0.95\columnwidth}{!}{%
\begin{tabular}{@{}c|c|c@{}}
\toprule
Method              & Encoder                                               & Decoder                                  \\ \midrule
MTrajRec            & $\text{O}(m \cdot d^2)$                               & \multirow{3}{*}{$\text{O}(m \cdot d^2)$} \\
RNTrajRec           & $\text{O}(m \cdot d^2 + m^2 \cdot d + \|G\| \cdot d^2)$ &                                          \\ 
\review{MM-STGED} & \review{$\text{O}(m^2 \cdot d^2)$ }\\
\midrule
\textbf{TedTrajRec} & $\text{O}(m^2 \cdot d^2)$                             & $\text{O}(m \cdot d^2)$                  \\ \bottomrule
\end{tabular}}
    
    \label{tab:timecost}
\end{table}


\review{Next, we conduct a comprehensive analysis of the computational overhead of the TedTrajRec framework using both static and runtime metrics:}
    \begin{itemize}
        \item \review{Parameters: The total number of trainable parameters in the model, indicating the overall model size.}
        \item \review{MACs (Multiply-Accumulate Operations): A critical metric for assessing computational complexity in neural networks, it measures the total number of multiply-accumulate operations performed by the model.}
        \item \review{Max Memory: The peak memory consumption during the training process, offering insight into the model's memory requirements.}
        \item \review{Epoch Time: The average time required to train the model per epoch with a dataset in Chengdu consisting of 105,879 trajectories.}
    \end{itemize}

\review{Table~\ref{tab: effi} presents an efficiency comparison of various map-constrained trajectory recovery models, including MTrajRec, RNTrajRec, MM-STGED, and TedTrajRec, evaluated on the Chengdu dataset. Among these models, TedTrajRec strikes the best balance between performance and efficiency. 
 RNTrajRec requires significantly 9.1 times more memory compared to MTrajRec while achieving a 4.21\% improvement in accuracy. Additionally, MM-STGED has 3.28\% lower computational cost (MACs) and 31.65\% lower memory usage compared to RNTrajrec, placing it between RNTrajRec and TedTrajRec in terms of resource efficiency. In terms of performance, MM-STGED slightly outperforms RNTrajRec but still significantly lags behind TedTrajRec.} 

\review{Furthermore, benefiting from the batching technique, in which trajectories are processed with a mini-batch of $64$ during the training process, the time cost for processing a single trajectory is $11.9$ milliseconds, making it highly suitable for real-world applications. Techniques such as minimizing the hidden dimension and reducing the number of Transformer layers could further reduce computational time to support various application scenarios.}


\subsection{Ablation Study}

\begin{table}[t]
\renewcommand\tabcolsep{3.3pt} 
\caption{Ablation study of different time-aware attention mechanisms in TedFormer.}
\vspace{-0.15in}
\begin{center}
\resizebox{\columnwidth}{!}{%
\begin{tabular}{c|cccccc}
\toprule
\multirow{2}{*}{Variants} &\multicolumn{6}{c}{Chengdu ($\epsilon_\rho=\epsilon_\tau*8$)} \\
& Recall & Precision & F1 Score & Accuracy & MAE & RMSE  \\
\midrule
Full Attention & 0.7782 & \underline{0.8918} & 0.8291 & 0.6578 & 137.42 & 226.20 \\
Exp. Decay & \underline{0.8166} & 0.8899 & \underline{0.8499} & \underline{0.7247} & \underline{120.68} & \underline{222.60} \\
\midrule
TedAttn & \textbf{0.8247} & \textbf{0.8936} & \textbf{0.8561} & \textbf{0.7301} & \textbf{108.33} & \textbf{192.22} \\
\bottomrule
\end{tabular}}

\vspace{5pt}

\resizebox{\columnwidth}{!}{%
\begin{tabular}{c|cccccc}
\toprule
\multirow{2}{*}{Variants} &\multicolumn{6}{c}{Porto ($\epsilon_\rho=\epsilon_\tau*8$)} \\
& Recall & Precision & F1 Score & Accuracy & MAE & RMSE  \\
\midrule
Full Attention & 0.6605 & \underline{0.8105} & 0.7253 & 0.5328 & 97.18 & 144.70 \\
Exp. Decay & \underline{0.6877} & 0.8083 & \underline{0.7407} & \underline{0.5665} & \underline{86.37} & \underline{133.15} \\
\midrule
TedAttn & \textbf{0.7097} & \textbf{0.8120} & \textbf{0.7554} & \textbf{0.5852} & \textbf{79.49} & \textbf{123.86} \\
\bottomrule
\end{tabular}}
\end{center}
\label{tab: ablation1}
\end{table}

\begin{table}[t]
\renewcommand\tabcolsep{3.3pt} 
\caption{Ablation study of different spatial and temporal methods in PD-GNN module.}
\vspace{-0.15in}
\begin{center}
\resizebox{\columnwidth}{!}{%
\begin{tabular}{c|cccccc}
\toprule
\multirow{2}{*}{Variants} &\multicolumn{6}{c}{Chengdu ($\epsilon_\rho=\epsilon_\tau*8$)} \\
& Recall & Precision & F1 Score & Accuracy & MAE & RMSE  \\
\midrule
w/o GNN & 0.8125 &0.8952 &0.8502 &0.7173 & 117.06 &202.76 \\
w/o time & \underline{0.8198}          & \textbf{0.8954} & \underline{0.8543}    & 0.7277          & \underline{112.54}    & \underline{198.41} \\
w/o dynamic & 0.8173    & \underline{0.8939}    & 0.8522          & \underline{0.7280}    & 112.61          & 201.02   \\
\midrule
PD-GNN & \textbf{0.8247} & 0.8936 & \textbf{0.8561} & \textbf{0.7301} & \textbf{108.33} & \textbf{192.22} \\
\bottomrule
\end{tabular}}

\vspace{5pt}

\resizebox{\columnwidth}{!}{%
\begin{tabular}{c|cccccc}
\toprule
\multirow{2}{*}{Variants} &\multicolumn{6}{c}{Porto ($\epsilon_\rho=\epsilon_\tau*8$)} \\
& Recall & Precision & F1 Score & Accuracy & MAE & RMSE  \\
\midrule
w/o GNN & 0.7028 &0.8110 & 0.7509 & 0.5765 & 82.93& 128.58 \\
w/o time & 0.7065 & \textbf{0.8133} & 0.7541 & 0.5783 & 81.70 & 125.19 \\
w/o dynamic & \underline{0.7085} & \underline{0.8129} & \underline{0.7551} & \underline{0.5827} & \underline{80.46} & \underline{124.57} \\ 
\midrule
PD-GNN & \textbf{0.7097} & 0.8120 & \textbf{0.7554} & \textbf{0.5852} & \textbf{79.49} & \textbf{123.86} \\
\bottomrule
\end{tabular}}
\end{center}
\label{tab: ablation2}
\end{table}

\begin{table}[t]
\renewcommand\tabcolsep{3.3pt} 
\caption{Ablation study of different graph neural networks in the spatial modeling of PD-GNN.
}
\vspace{-0.15in}
\begin{center}
\resizebox{\columnwidth}{!}{%
\begin{tabular}{c|cccccc}
\toprule
\multirow{2}{*}{Variants} &\multicolumn{6}{c}{Chengdu ($\epsilon_\rho=\epsilon_\tau*8$)} \\
& Recall & Precision & F1 Score & Accuracy & MAE & RMSE  \\
\midrule
w/o GNN & 0.8125 &0.8952 &0.8502 &0.7173 & 117.06 & 202.76 \\
\midrule
GCN & 0.8147  & 0.8949 & 0.8512   & 0.7286         & 109.96    & 192.61 \\
GraphSage & 0.8156          & \textbf{0.8967} & 0.8526    & 0.7304 & 110.55    & 195.57 \\
GIN & \underline{0.8211}  & \underline{0.8960} & \underline{0.8552}    & 0.7303          & 109.90    & \underline{193.76} \\
GAT & 0.8212    & 0.8954    & 0.8550          & \textbf{0.7317}   & \underline{109.57}         & 195.10  \\
\midrule
GATv2 & \textbf{0.8247} & 0.8936 & \textbf{0.8561} & \underline{0.7301} & \textbf{108.33} & \textbf{192.22} \\
\bottomrule
\end{tabular}}

\vspace{5pt}

\resizebox{\columnwidth}{!}{%
\begin{tabular}{c|cccccc}
\toprule
\multirow{2}{*}{Variants} &\multicolumn{6}{c}{Porto ($\epsilon_\rho=\epsilon_\tau*8$)} \\
& Recall & Precision & F1 Score & Accuracy & MAE & RMSE  \\
\midrule
w/o GNN & 0.7028 &0.8110 & 0.7509 & 0.5765 & 82.93& 128.58 \\
\midrule
GCN & 0.7051 & \underline{0.8159} & 0.7544 & 0.5823 & 81.17 & 125.63 \\
GraphSage & 0.7055 & 0.8140 & 0.7538 & 0.5835 & 80.50 & 124.85 \\ 
GIN & \textbf{0.7102} & 0.8114 & \textbf{0.7554} & 0.5845 & \underline{79.63} & \textbf{123.31} \\ 
GAT & 0.7050 & \textbf{0.8174} & 0.7550 & \textbf{0.5867} & 79.87 & 123.52 \\ 
\midrule
GATv2 & \underline{0.7097} & 0.8120 & \textbf{0.7554} & \underline{0.5852} & \textbf{79.49} & \underline{123.86} \\
\bottomrule
\end{tabular}}
\end{center}
\label{tab: ablation3}
\vspace{-0.1in}
\end{table}

The main components and contributions of our work include TedFormer, a time-aware Transformer that integrates spatio-temporal trajectory dynamics into the attention mechanism, and PD-GNN, a dynamic-aware road network representation module. 
To comprehensively demonstrate the effectiveness of these components, we conduct 
ablation experiments. 

Firstly, we conduct ablation experiments on different attention mechanisms. In particular, we replace \textbf{TedAttn} defined in Eq.~\eqref{eq: tedattn} with:
\begin{itemize}
    \item \textbf{Full Attention}~\cite{vaswani2017attention}, which adopts a vanilla self-attention in Transformer architecture;
    \item \textbf{Exponential Decay Attention}~\cite{jiang2023self}, which utilizes a log-scaled exponential decay function to include time interval information.
\end{itemize}

Experimental results on the Chengdu and Porto datasets, reported in Table~\ref{tab: ablation1}, demonstrate that TedAttn consistently outperforms other attention mechanisms on both datasets. Full attention yields the worst results, emphasizing the importance of incorporating time information into the attention mechanism for the trajectory recovery task. Moreover, while the Exponential Decay attention mechanism captures temporal information, it lacks the ability to capture complex dynamics of trajectory evolution. \review{Furthermore, TedFormer simultaneously addresses both irregular time interval information and spatial-temporal trajectory dynamics, leading to superior performance compared to existing trajectory representation approaches.}

Next, we study the impact of periodic dynamic-aware road network representation and graph neural networks. Specifically, we replace Eq.~\eqref{Eq:road embedding} with: 

\begin{figure*}[t]
    \centering
    \includegraphics[width=\linewidth]{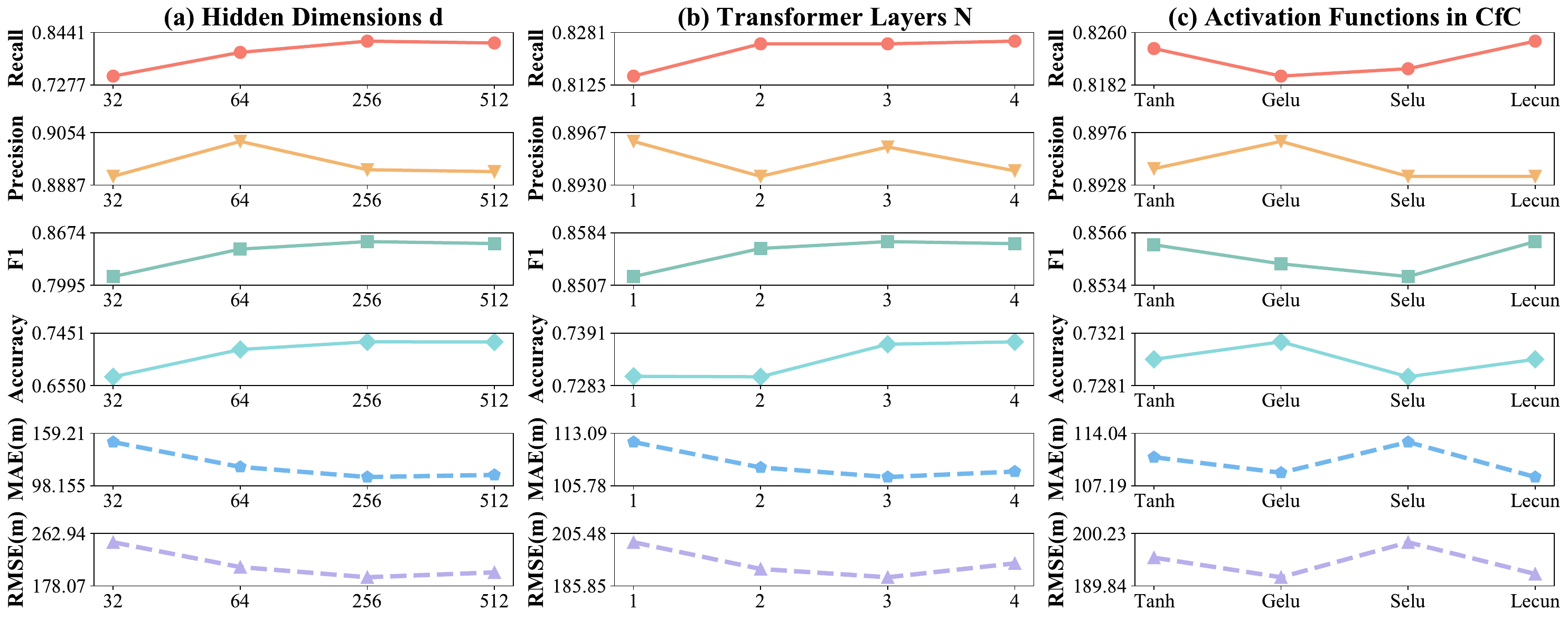}
    \caption{The Performance of TedTrajRec on Chengdu Dataset ($\epsilon_\rho=\epsilon_\tau * 8$)
under Different Hyper-Parameters. The solid lines indicate that higher metric values are preferable, whereas the dashed lines signify that lower values are more desirable.}
    \label{fig:parameter1}
\end{figure*}
\begin{figure*}[t]
    \centering
    \includegraphics[width=\linewidth]{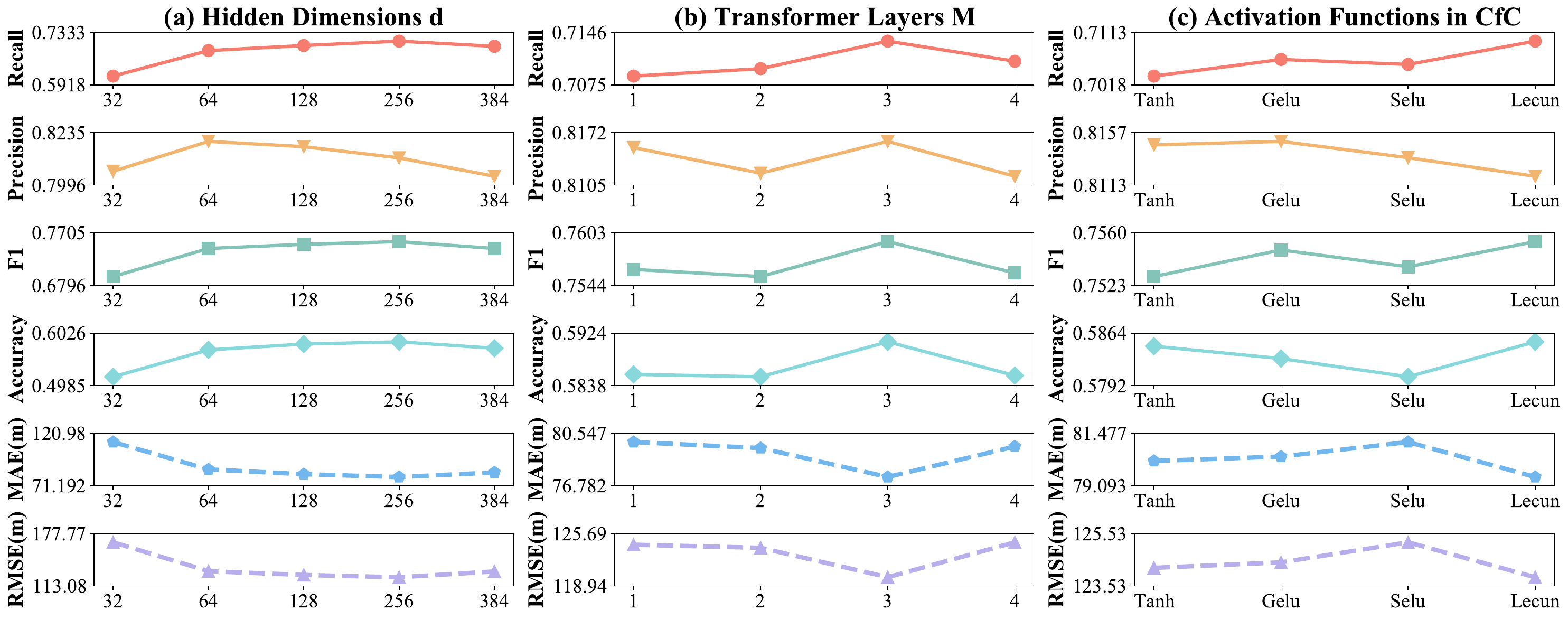}
    \caption{The Performance of TedTrajRec on Porto Dataset ($\epsilon_\rho=\epsilon_\tau * 8$)
under Different Hyper-Parameters. The solid lines indicate that higher metric values are preferable, whereas the dashed lines signify that lower values are more desirable.}
    \label{fig:parameter2}
\end{figure*}

\begin{itemize}
    \item \textbf{w/o GNN}, i.e., we replace the GATv2 module in Eq.~\eqref{eq:gat-attn} with the linear network, resulting in $X_{e_i}^{road}(t) = \textbf{t2v}(t; \Omega_i) + S_i$;
    \item \textbf{w/o time}, i.e., $X_{e_i}^{road} = \hat{S}_i$, that does not consider time information;
    \item \textbf{w/o dynamic}, i.e., $X_{e_i}^{road}(t) = \textbf{t2v}(t; \omega)+\hat{S}_i$, where $\omega \in \mathbb{R}^{d}$ remains constant for each road segment.
\end{itemize}

Experimental results in Table~\ref{tab: ablation2} highlight the importance of integrating spatial and temporal dynamics. \review{Furthermore, we observe that when $\omega_i$ is set to all zeros, PD-GNN degrades to the variant of \textbf{w/o time}, which corresponds to the road network representation module used in RNTrajRec. This is currently the only approach specifically designed for map-constrained trajectory recovery that captures the spatial topology of the road network. In contrast, PD-GNN provides a more robust representation capability by integrating both spatial and temporal aspects.}

Additionally, we examine the performance of different graph neural networks within the PD-GNN module. Specifically, we adopt GATv2 in Eq.~\eqref{eq:gat-attn} for spatial modeling. We replace GATv2 with:
\begin{itemize}
    \item \textbf{GCN~\cite{kipf2016semi}}, non-attentive graph neural networks that aggregate node features based on local graph structure.
    \item \textbf{GraphSage~\cite{hamilton2017inductive}},  inductive non-attentive graph neural networks that generate node embeddings by sampling and aggregating features.
    \item \textbf{GIN~\cite{xu2018powerful}}, non-attentive graph neural networks designed to be as powerful as the Weisfeiler-Lehman graph isomorphism.
    \item \textbf{GAT~\cite{velivckovic2017graph}} graph attention networks that weigh the importance of neighboring nodes during feature aggregation. 
\end{itemize}

As depicted in Table~\ref{tab: ablation3}, GATv2 excels in MAE, RMSE, and F1 Score, while GAT leads in Accuracy.
Notably, we can observe that attention-based methods outperform non-attention methods.


\begin{figure*}[t]
    \centering
    \includegraphics[width=1.0\linewidth]{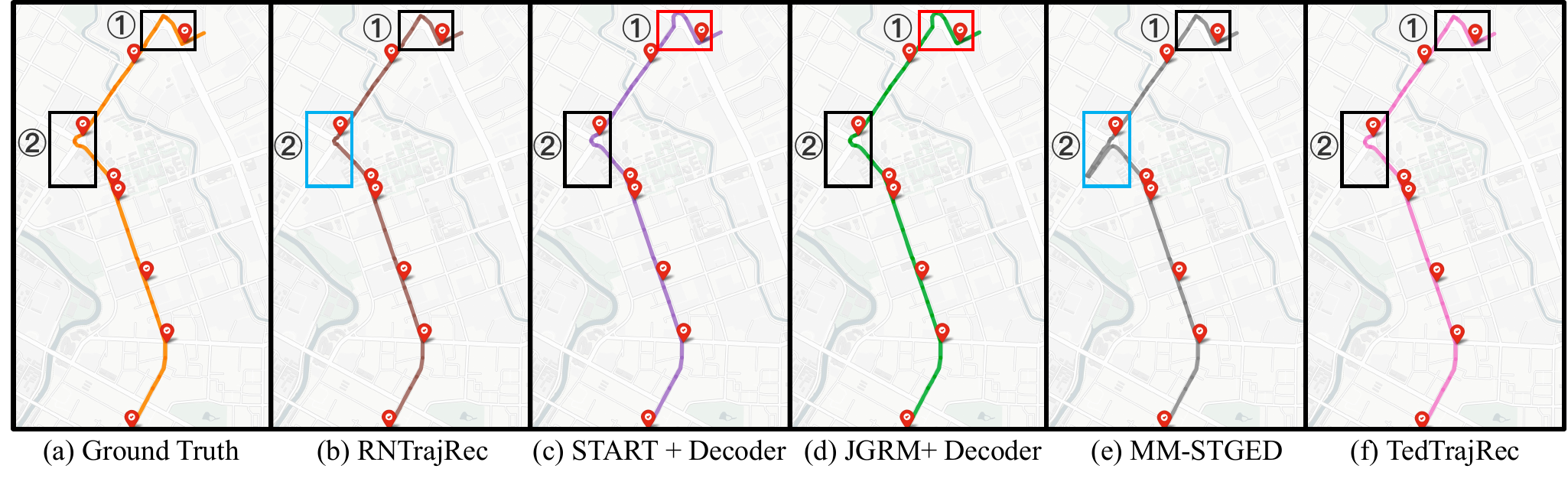}
    \caption{Case Study of Trajectory Recovery on Chengdu Dataset ($\epsilon_\rho=\epsilon_\tau * 8$).  
    }
    \label{fig:case_study}
    \vspace{-0.1in}
\end{figure*}

\subsection{Parameter Analysis}

In the following series of experiments, we evaluate the impacts of three key parameters: the number $M$ of encoder layers, the dimensionality $d$ of hidden-state vectors, and the choice of activation function in CfC cell. We conduct the experiments on Chengdu and Porto datasets, as shown in Figure~\ref{fig:parameter1} and Figure~\ref{fig:parameter2}, respectively.

\noindent \textbf{Influence of the dimensionality of hidden-state vectors $d$.} 
The size $d$ of hidden state vectors is a crucial factor affecting TedTrajRec's complexity. Our exploration, as depicted in Figures~\ref{fig:parameter1}(a) and \ref{fig:parameter2}(a), varies $d$ from $32$ to $512$ for the Chengdu dataset and $32$ to $384$ for the Porto dataset due to memory limitation. The results reveal that TedFormer attains optimal accuracy at $d=256$, which we adopt for the entire experiment.

\noindent \textbf{Influence of TedFormer encoder layers $M$.} 
The number of TedFormer encoder layers $M$ impacts the complexity of TedTrajRec. We analyze its effect by varying 
$M$ from $1$ to $4$ and report the results in Figures~\ref{fig:parameter1}(b) and \ref{fig:parameter2}(b). An excessive $M$ can lead to overfitting and reduced performance. We observe that TedFormer achieves the highest accuracy when $M=3$. For efficiency and GPU memory considerations, we opt for $M=2$.

\noindent \textbf{Influence of activation function in CfC cell.} 
Previous study \cite{kidger2020neural} highlights the potential influence of the activation function on Neural ODEs. To further examine this impact, we compare \textit{Lecun} activation function with \textit{Tanh}, \textit{GeLU}, and \textit{SeLU} activation functions and list the results in Figures~\ref{fig:parameter1}(c) and \ref{fig:parameter2}(c). Experimental results indicate that TedFormer shows limited sensitivity to the choice of activation function. 

\begin{table}[t]
\renewcommand\tabcolsep{3.3pt} 
\caption{Experimental results on upstream tasks, including trajectory similarity computation and trajectory clustering. Higher value for all metrics indicates a better performance.}
\begin{center}
\resizebox{\columnwidth}{!}{%
\begin{tabular}{c|cccccc}
\toprule
& \multicolumn{6}{c}{Chengdu ($\epsilon_\rho=\epsilon_\tau*8$) with \textbf{LCSS} metric} \\
\multirow{2}{*}{Method} & \multicolumn{3}{c}{Trajectory Similarity} & \multicolumn{3}{c}{Trajectory Clustering} \\
\cmidrule(lr){2-4}
\cmidrule(lr){5-7}
& R@1 & R@10 & R@50 & Homo. & Comp. & AdR.  \\
\midrule
Raw Data & 0.0814 & 0.3394 & 0.6308 & 0.2027 & 0.2683 & 0.2771 \\
\midrule
Linear + HMM & 0.1811 & 0.6090 & 0.8814 & 0.2962 & 0.3271 & 0.3856 \\
TrImpute + HMM &0.1747 & 0.5944& 0.8653& 0.2825&0.3514  &0.3832  \\
DHTR + HMM & 0.1438 & 0.5418 & 0.8526 & 0.3391 & 0.3464 & 0.4227 \\
TERI + HMM &0.2039 &0.6679 & 0.9126&0.3227 &  0.3727&  0.4100\\
\midrule

NeuTraj + Decoder & 0.2326 & 0.7196 & 0.9338 & 0.4463 & 0.4218 & 0.4732 \\
GTS + Decoder & 0.2409 & 0.7220 & 0.9387 & 0.4650 & 0.4313 & 0.4937 \\
TrajGAT + Decoder & 0.2312 & 0.7134 & 0.9374 & 0.4518 & 0.4390 & 0.4874 \\
TrajFormer + Decoder &0.2393  &0.7271  &0.9427 & 0.4435  & 0.4358  & 0.4903  \\
\review{START + Decoder} & \review{0.2523} & \review{0.7509} & \review{\underline{0.9516}} & \review{\underline{0.4876}} & \review{0.4575} & \review{0.5357}\\
\review{JGRM + Decoder} & \review{0.2506} & \review{0.7397} & \review{0.9445} & \review{0.4517}& \review{0.4320} & \review{0.4915}\\
MTrajRec & 0.2274 & 0.7035 & 0.9296 & 0.4325 & 0.4103 & 0.4580 \\
RNTrajRec & \underline{0.2563} & \underline{0.7515} & {0.9499} & {0.4858} & \underline{0.4756} & 0.5296 \\
\review{MM-STGED} &  \review{0.2558} & \review{0.7506} &  \review{0.9438} &  \review{0.4848} & \review{0.4653}&  \review{\underline{0.5422}}  \\
\textbf{TedTrajRec} & \textbf{0.2864} & \textbf{0.7784} & \textbf{0.9597} & \textbf{0.5339} & \textbf{0.5053} & \textbf{0.5998} \\
\bottomrule
\end{tabular}}

\vspace{5pt}

\resizebox{\columnwidth}{!}{%
\begin{tabular}{c|cccccc}
\toprule
& \multicolumn{6}{c}{Chengdu ($\epsilon_\rho=\epsilon_\tau*8$) with \textbf{EDR} metric} \\
\multirow{2}{*}{Method} & \multicolumn{3}{c}{Trajectory Similarity} & \multicolumn{3}{c}{Trajectory Clustering} \\
\cmidrule(lr){2-4}
\cmidrule(lr){5-7}
& R@1 & R@10 & R@50 & Homo. & Comp. & AdR.  \\
\midrule
Raw Data & 0.0754 & 0.3096 & 0.5943 & 0.1742 & 0.1846 & 0.2127 \\
\midrule
Linear + HMM & 0.1580 & 0.5597 & 0.8403 & 0.3194 & 0.3471 & 0.3971 \\
TrImpute + HMM &0.1581 &0.5495 &0.8297 &0.3176 &0.3441  & 0.3998 \\
DHTR + HMM & 0.1277 & 0.4884 & 0.8022 & 0.3167 & 0.3521 & 0.4253 \\
TERI + HMM & 0.1762&0.6130 &0.8730 &0.3671 & 0.3654 & 0.4441 \\
\midrule

NeuTraj + Decoder & 0.2063 & 0.6565 & 0.8986 & 0.4299 & 0.4354 & 0.5181 \\
GTS + Decoder & 0.2108 & 0.6579 & 0.9035 & 0.4351 & 0.4317 & 0.5219  \\
TrajGAT + Decoder & 0.2041 & 0.6458 & 0.8980 & 0.4275 & 0.4314 & 0.5262 \\
TrajFormer + Decoder &0.2122  & 0.6609 &0.9039 & 0.4204  &0.4407   &0.5293   \\

\review{START + Decoder} & \review{0.2211} & \review{0.6789} & \review{0.9167} & \review{0.4570} & \review{\underline{0.4721}} & \review{0.5694}\\
\review{JGRM + Decoder} & \review{0.2185} & \review{0.6749} & \review{0.9119} & \review{0.4380} & \review{0.4665} & \review{0.5505}\\
MTrajRec & 0.1958 & 0.6387 & 0.8914 & 0.4225 & 0.4285 & 0.5109  \\
RNTrajRec & 0.2246 & \underline{0.6887} & \underline{0.9176} & \underline{0.4700} & {0.4678} & \underline{0.5726} \\
\review{MM-STGED} & \review{\underline{0.2261}} & \review{0.6857} & \review{0.9161} & \review{0.4573} & \review{0.4553} & \review{0.5592}\\
\textbf{TedTrajRec} & \textbf{0.2482} & \textbf{0.7206} & \textbf{0.9314} & \textbf{0.4992} & \textbf{0.4892} & \textbf{0.5999} \\
\bottomrule
\end{tabular}}

\end{center}
\label{tab: sim}
\end{table}

\subsection{Case Study}

\review{We present a case study, illustrated in Figure~\ref{fig:case_study}, to highlight the importance of learning spatio-temporal dynamics for trajectory recovery. As shown in Figure~\ref{fig:case_study}(a), the trajectory from the Chengdu test dataset is depicted, with eight input GPS points marked in red. We compare the trajectory prediction results of TedTrajRec against four strong baselines in Chengdu: RNTrajRec, START, JGRM, and MM-STGED, as shown in Figures~\ref{fig:case_study}(b-f).}

\review{In the figure, while all five models approximate the high-sampling trajectory, our model’s recovery aligns most closely with the ground truth. We highlight two areas of the trajectory where the models show differing predictions. The black rectangle indicates where the prediction matches the ground truth, while the red and blue rectangles mark areas of deviation — red for area \textcircled{1} and blue for area \textcircled{2}. We observe that RNTrajRec and MM-STGED succeed in area \textcircled{1}, while START and JGRM obtain the correct prediction in area \textcircled{2}. Only TedTrajRec succeeds in both areas. Our model's precision in recovering the trajectory seamlessly from start to finish is strengthened by its ability to comprehensively model spatio-temporal dynamics, both within the road network and the trajectory itself, contributing to a more accurate representation of trajectory features.}


\subsection{Upstream Task Analysis}

Trajectory recovery aims at reinstating missing GPS points in low-sampling-rate GPS trajectories, enhancing the quality of trajectory data for upstream tasks. We specifically focus on two pivotal upstream tasks for in-depth analysis: trajectory similarity computation~\cite{zhang2020trajectory, han2021graph, yao2019computing} and trajectory clustering~\cite{yao2022trajgat}.
Given a set of trajectories $D^t$ and a similarity metric $\mathcal{M}$ (e.g., LCSS, DTW, ERP, etc), 
the former identifies the most similar trajectory for each $\tau \in D^t$, while the latter
evaluates clustering performance between similarity matrices. For simplicity, we normalize the similarity metric, assuming $\mathcal{M}(\cdot, \cdot) \in [0,1]$, where $\mathcal{M}(\tau, \tau^{'})=0$ only if $\tau=\tau^{'}$.

Assuming access to a trajectory dataset $D^t$ with a high sampling rate, where $\tau$ and $TS(\tau)$ represent a trajectory in $D^t$ and its true most similar trajectory, respectively, i.e.,
\begin{equation}
TS(\tau) = \arg \min\nolimits_{\tau^{'} \in D^t, \tau \neq \tau_{'}} \mathcal{M} (\tau, \tau^{'}) ~.
\end{equation}
Let $\rho=DS(\tau)$ be a low-sampling-rate trajectory down-sampled from $\tau \in D^t$, and $f(\cdot)$ represent a trajectory recovery model transforming a low-sampling-rate trajectory $\rho$ into a trajectory $f(\rho)$ with a higher sampling rate. We perform the trajectory similarity computation task on recovered trajectories $\cup f(\rho)$, using the evaluation metric R@k: 
\begin{equation}
    \text{R@k}=\frac{1}{|D^t|} \mathbf{1} \{ TS(\tau) \in \text{Topk}_{\tau^{'} \in D^t, \tau \neq \tau_{'}} \mathcal{M}(f(\rho), f(\rho^{'}))  \} ~,
\end{equation}
where Topk returns $k$ trajectories with the shortest distances to $f(\rho)$. Note $f(\tau)=\tau$ refers to the baseline that uses the low-sampling-rate trajectory directly for retrieval.

Additionally, considering the true distance matrix $\mathcal{D} \in \mathbb{R}^{|D^t| \times |D^t|}$, where $\mathcal{D}_{i,j}=1-\mathcal{M}(\tau_i, \tau_j)$, 
we define the recovered distance matrix as $\hat{\mathcal{D}}^f$, i.e.,
\begin{equation}
\hat{\mathcal{D}}^f=\hat{\mathcal{D}}^f_{i,j}=1-\mathcal{M}(f(DS(\tau_i)), f(DS(\tau_j))) ~.
\end{equation}
Clustering similarity is measured between $\mathcal{D}$ and $\hat{\mathcal{D}}^f$ using homogeneity score (Homo.), completeness score (Comp.), and adjusted rand score (AdR.). Notably, these metrics are available in \textit{scikit-learn}. Following the experimental protocol from \cite{yao2022trajgat}, we adopt the DBSCAN clustering algorithm and report the averaged score over different $\epsilon$ values. Notably, we search $\epsilon$ within sample pairs of one cluster in the range of $[0.45, 0.55]$, as both values $>0.55$ and values $<0.45$ result in a minimal number of ($<10)$ large clusters. 

We focus on the low-sampling-rate trajectory similarity computation and trajectory clustering in spatial network~\cite{fang2022spatio}, using the longest common subsequence (LCSS) and edit distance on real sequence (EDR) metrics. Specifically, we selected $15,000$ trajectories from the Chengdu testing dataset to conduct the experiments. The experimental results, reported in Table~\ref{tab: sim}, demonstrate that TedTrajRec consistently outperforms its competitors. This underscores our model's ability to recover more information and highlights the improved accuracy of low-sampling-rate trajectory similarity retrieval through trajectory recovery, affirming its capability to restore lost information in low-sampling-rate trajectories. \review{For example, TedTrajRec improves 10.82\% on R@1 metrics and 7.62\% on the adjusted rand score metric, as compared to the second-best performers on the LCSS metric.} We also observe that the Raw Data baseline using low-sampling-rate trajectories directly (i.e., $f(\tau)=\tau$) performs the worst, showcasing that low-sampling-rate trajectories lose significant information. 



\section{Limitation and Future Work}

\review{TedTrajRec demonstrates its effectiveness and potential application values in the fields of trajectory recovery for low-sample GPS trajectory. However, there are several potential limitations of TedTrajRec that warrant discussion here:}
\begin{itemize}
    \item \review{
Limited Generalization for Cross-City Trajectory Recovery: The road networks in different cities can vary significantly. As a result, a model trained on data from one city may struggle to generalize effectively to another, leading to notable performance degradation.}
    \item \review{
Need for Post-Processing in Practical Applications: Although TedTrajRec takes road network structure into account, the recovered trajectories may still be unrealistic due to neural networks' inherent uncertainty. 
Therefore, a post-processing algorithm or carefully designed decoding constraints are necessary to ensure practical applicability.}
    \item \review{
Memory Constraints for Large-Scale Training: The memory requirements of TedTrajRec are closely tied to the size of the road network, specifically the number of road segments within the training area. As a result, training on cities or countries with millions of road segments can lead to memory constraints. To address this, it may be necessary to divide larger regions into smaller, more manageable training areas.}
\end{itemize}

\review{For future work, we intend to explore domain adaptation techniques to improve the model's transferability across different cities. Additionally, we aim to incorporate real-time traffic data and enhance the recovery model's constraints to improve its robustness and applicability. Furthermore, we will investigate strategies to address memory challenges when scaling to larger regions, such as hierarchical modeling or distributed training approaches.}

\section{Conclusion}

We propose TedTrajRec, a spatio-temporal dynamic-aware Transformer architecture with a time-aware Transformer model that incorporates continuous-time dynamics into the attention model of Transformer architecture, and a periodic dynamic-aware graph neural network that captures spatio-temporal patterns of the road network.
Extensive experiments on three real-world datasets demonstrate the superiority of the proposed model and its practical impact. 


\bibliographystyle{IEEEtran}
\bibliography{reference}

\vspace{11pt}

\begin{IEEEbiography}
[{\includegraphics[width=1in,height=1.25in, clip,keepaspectratio]{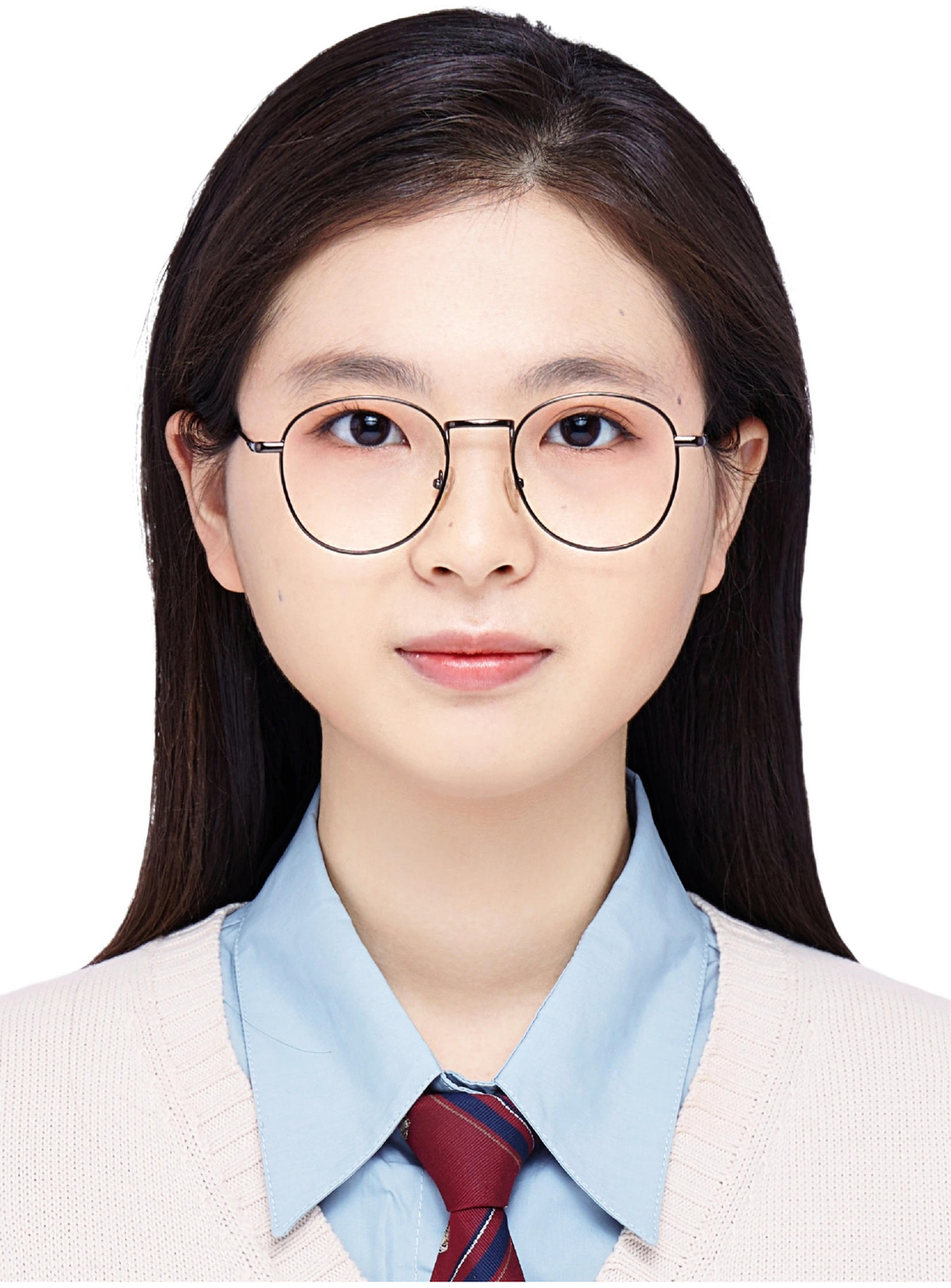}}]{Tian Sun} received the B.E. degree in software engineering from East China Normal University, China, in 2023. She is currently pursuing the master’s degree in computer science and technology from Fudan University, China. Her research interests include trajectory recovery and spatio-temporal data analysis.
\end{IEEEbiography}

\begin{IEEEbiography}
[{\includegraphics[width=1in,height=1.25in, clip,keepaspectratio]{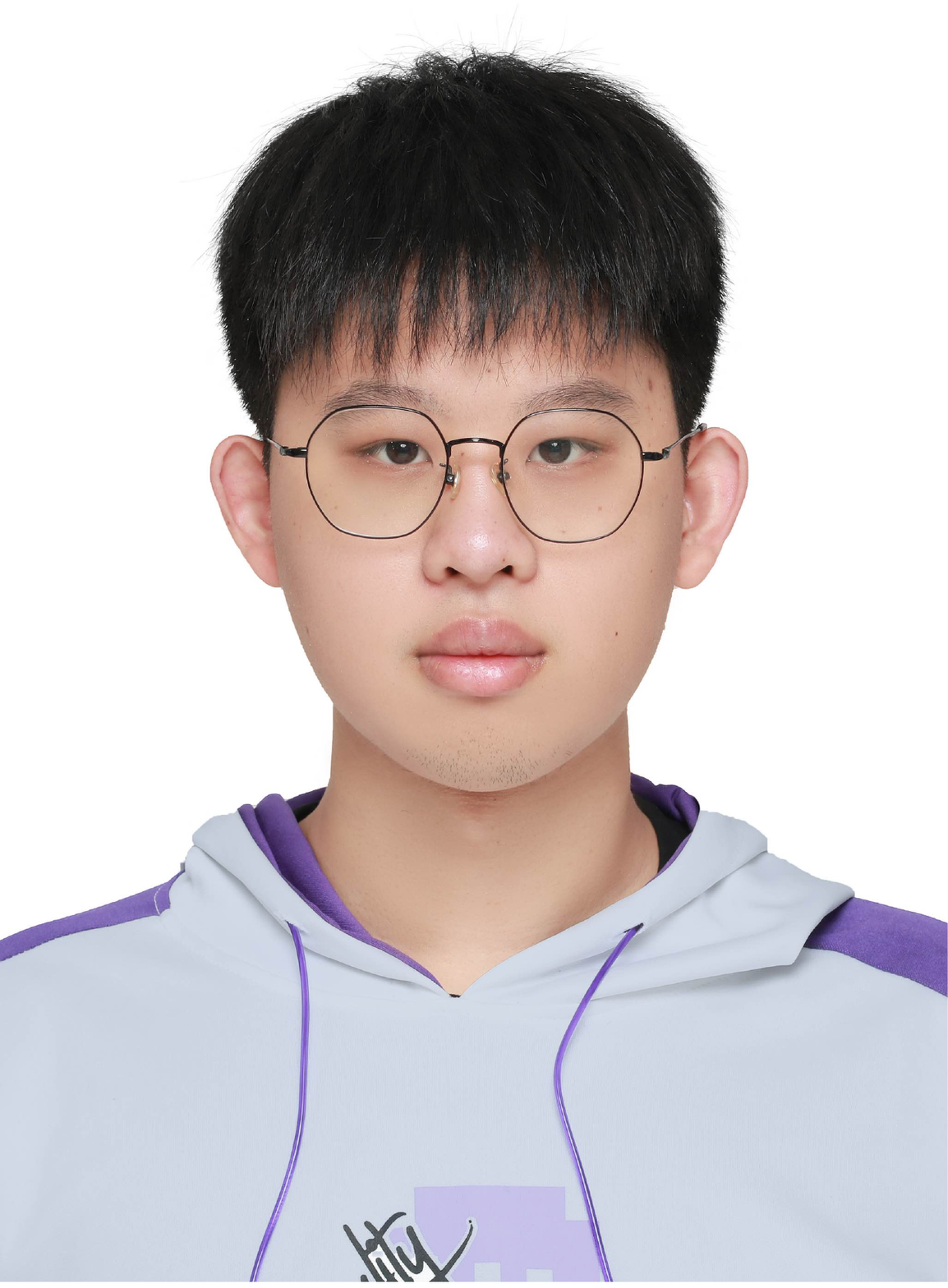}}]{Yuqi Chen} received a master's degree in computer science from Fudan University, China, in 2023. He is currently working at Xiaohongshu Inc. His research interests include trajectory recovery, spatio-temporal data analysis, and time-series modeling. Work was done during his purchasing of master's degree.
\end{IEEEbiography}

\begin{IEEEbiography}
[{\includegraphics[width=1in,height=1.25in,clip,keepaspectratio]{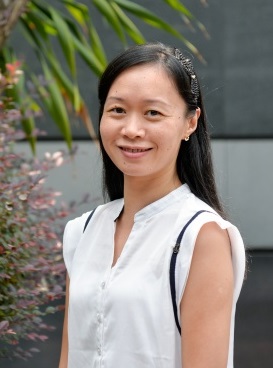}}]{Baihua Zheng} received the Ph.D. degree from
the Department of Computer Science, Hong Kong
University of Science and Technology in 2003. She
is currently a professor at the School of Computing and Information Systems, at Singapore Management University. Her research interests include mobile/pervasive computing, spatial databases, and big data analytics.
\end{IEEEbiography}

\vspace{0.3in}

\begin{IEEEbiography}
[{\includegraphics[width=1in,height=1.25in,clip,keepaspectratio]{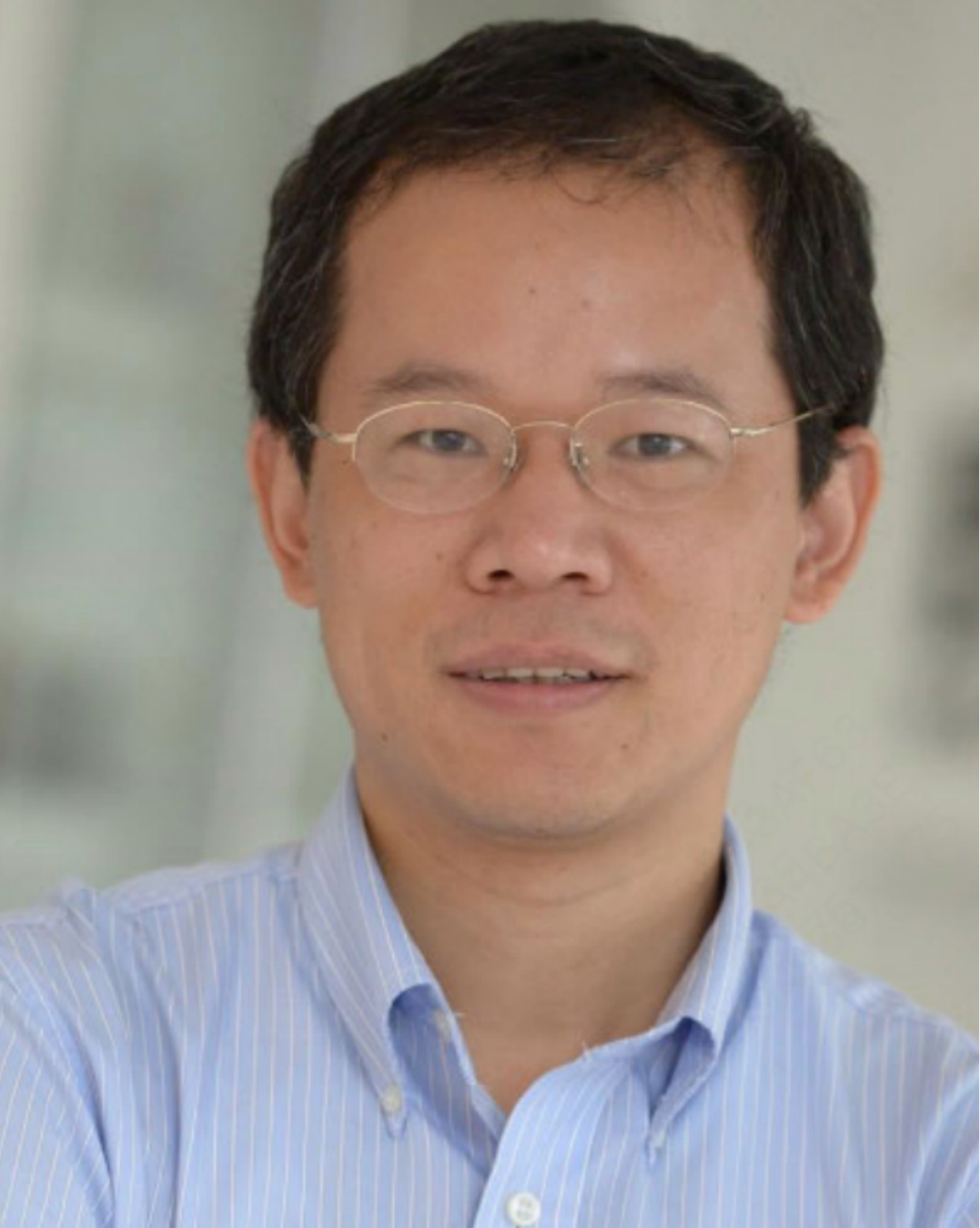}}]{Weiwei Sun} received a Ph.D. degree in computer software and theory from Fudan University, China, in 2002. He is currently a Professor at the School of Computer Science, Fudan University. His research interests include big data, data mining, intelligent transportation and logistics, deep learning, reinforcement learning, and their applications in traffic signal control, spatio-temporal data analysis, and trajectory pretraining.
\end{IEEEbiography}

\vspace{11pt}

\vfill

\end{document}